%% file: neurips_2026.tex
\documentclass{article}

 \usepackage[preprint]{neurips_2026}

\usepackage[utf8]{inputenc} 
\usepackage[T1]{fontenc}    
\usepackage{hyperref}       
\usepackage{url}            
\usepackage{booktabs}       
\usepackage{amsfonts}       
\usepackage{nicefrac}       
\usepackage{microtype}      
\usepackage{xcolor}         
\usepackage{graphicx}
\usepackage{wrapfig}
\usepackage{subcaption}
\usepackage{booktabs}
\usepackage{multirow}
\usepackage{makecell}

\title{\textsc{HyperZip}: Efficient Data Compression through Personalized Diffusion LLMs with Hypernetworks.}

\author{%
  Thai Nguyen \qquad Khang Tran \qquad NhatHai Phan \\
  New Jersey Institute of Technology \\
  \texttt{\{tqn,kt36,phan\}@njit.edu}
}

\begin{document}

\maketitle

\begin{abstract}
    Large language models (LLMs) have shown strong potential for lossless data compression, but existing approaches are constrained by the high computational cost and low throughput of autoregressive decoding. We propose \textsc{\textbf{HyperZip}}, an efficient and scalable LLM-based compression framework that leverages diffusion-based LLMs (dLLMs) with Multi-Token Prediction (MTP) to accelerate LLM-based data compression processes. We identify a trade-off in diffusion-based compression, where increasing decoding throughput degrades the compression rate. To mitigate this trade-off, \textsc{HyperZip} employs a hypernetwork to generate data-specific updates from a context representation, adapting the dLLM to the target data without costly fine-tuning, resulting in a low compression rate and high throughput. Extensive experiments show that \textsc{HyperZip} achieves a superior trade-off between compression rate and speed compared with state-of-the-art baselines.
\end{abstract}

\section{Introduction}

\textbf{Motivation.} The advances in LLMs have been adopted in many real-world applications across multiple domains~\cite{raiaan2024review, hagos2024recent, chkirbene2024large, naveed2025comprehensive}, including data management and storage~\cite{miao2024demystifying, zhang2024applications, bagui2026large, tummuri2025enhancing}. Recent works show that LLM-based data compressors can achieve better compression rates than traditional methods such as gzip, resulting in more efficient data storage \cite{deletang2024language,li2025lossless,chen2026large, pan2025understanding}. However, existing LLM-based data compressors are limited by the resources required to deploy an LLM and its autoregressive decoding generative mechanism, which requires high-performance servers to execute the compressing and decompressing tasks ~\cite{kipf2026waiting, haddadi2026comparing}. This degrades compression and decompression speed and limits adoption in real-world use, where users only have limited computing resources. Therefore, it's critical to develop a scalable, efficient LLM-based data compressor that compresses quickly while maintaining a low compression rate.

\textbf{Challenges.} Designing such a mechanism is very challenging. Because existing LLM-based data compressors use LLMs as the probabilistic model for lossless arithmetic encoding, they face the following challenges. \textbf{(1)} The typical generation process of LLMs follows the autoregressive generation pipeline, which only generates the logit for a single token in one forward pass, limiting the compression speed of large files with millions of tokens~\cite{kipf2026waiting}. \textbf{(2)} To achieve a lower compression rate, the LLMs used for compression must exhibit low perplexity on the data being compressed, thereby assigning higher confidence to the predicted logits of subsequent tokens. Achieving such low perplexity typically requires either complex models with billions of parameters or an additional fine-tuning process tailored to the target data~\cite{zhang2025l3tc}.

\textbf{Our solution.} In this work, we propose \textsc{\textbf{HyperZip}}, a novel, efficient, and scalable framework of LLM-based data compressors. \textsc{HyperZip} addresses the first challenge by replacing autoregressive LLMs with diffusion-based LLMs (dLLMs) that can perform Multi-Token Prediction (MTP), effectively increasing compression speed. Our preliminary results (details in Appendix \ref{app:mtp-study}) show that diffusion-based MTP can significantly improve compression speed while maintaining a low compression rate, outperforming other existing MTP mechanisms~\cite{gloeckle2024better, hu2025speculative}. However, our results also reveal a trade-off between compression ratio and speed when using diffusion-based MTP for data compression. Specifically, we can achieve faster compression by lowering the confidence threshold for diffusion decoding, which increases perplexity on the compressed data and yields a higher compression rate. This behavior reflects the throughput-performance trade-off inherent in diffusion-based LLMs~\cite{fu2025bits}. As with the second challenge, more complex, generalized models or an additional fine-tuning step can address this, but they require more computational resources. 

To address this issue, the core of \textsc{HyperZip} employs a hypernetwork pipeline~\cite{chauhan2024brief, tan2026instant} to generate a LoRA update~\cite{hu2021lora} for the dLLMs given a context vector representing the data to be compressed, adapting the model's weights to the data. This adaptation improves the model's predictive confidence and reduces the compression rate without requiring costly additional fine-tuning. At inference time, users can compress the context vector along with the target data, which the model then uses to update during decompression. This pipeline lets \textsc{HyperZip} personalize dLLMs to the data without expensive fine-tuning while maintaining low compression rates. As a result, \textsc{HyperZip} achieves the best trade-off between compression rate and speed, facilitating practical adoption.

\textbf{Contribution.} Our contribution is as follows: \textbf{(1)} We propose \textsc{HyperZip}, an efficient and scalable framework for LLM-based data compression. \textsc{HyperZip} replaces conventional LLMs with dLLMs that can perform multi-token prediction, enabling faster compression for real-world adoption. \textbf{(2)} To address the trade-off between compression rate and speed of dLLMs, we design a Hypernetwork structure to personalize diffusion models, aligning them to the data, desired compression speed, and latency. This makes LLM-based data compression practical for real-world use. \textbf{(3)} We conduct extensive experiments to evaluate \textsc{HyperZip}, and the results show that \textsc{HyperZip} achieves the best trade-off between compression rate and speed compared with state-of-the-art baselines.

\section{Background \& Preliminaries}
\textbf{Arithmetic Coding.} Given a data file which contain a sequence $X = [x_1, x_2,...,x_n] \in V^n$ of length $n$ from a finite set of vocabulary $V$, arithmetic coding~\cite{pasco1977source, rissanen1976generalized} encodes a sequence $x_{1:n}$ into a floating number using a probability model $\rho$. It iteratively narrows an interval $I_k, k \in [n]$, starting from $I_0 = [0,1)$, by partitioning the interval into sub-intervals proportional to $\rho(x_k | x_{1:k-1})$, and keeping $I$ as the one corresponding to the interval associated with $x_k$. After $n$ steps, the final interval $I_n$ uniquely identifies $x_{1:n}$, and a floating number picked from $I_n$ is the compressed file.

\textbf{Neural compression.} Neural compression~\cite{yang2023introduction} employs an autoregressive language model as an approximation for the probability model $\rho$ in arithmetic coding~\cite{deletang2024language,li2025lossless,mittu2024finezip,heurteldepeiges2024compression,zhang2025testtime,pan2025understanding, bellard2021nncp}. At each encoding iteration $k \in [n]$, the arithmetic encoder replaces the probability model $\rho(x_k | x_{1:k-1})$ with an LLM's logit $q_\theta(x_k | x_{1:k-1})$, where $\theta$ is the LLM's parameters. These approaches have outperformed strong classical compressors, such as Gzip~\cite{deutsch1996gzip}, because LLMs generalize well and can accurately approximate a probability model $\rho$ of the target file. In addition, these methods are also applied efficiently in other domains, such as image and audio~\cite{deletang2024language, zhao2026omnizip, chen2025large, li2025lossless}.

\textbf{Multi-token prediction with dLLMs}~\cite{austin2021structured,hoogeboom2021argmax, shi2024simplified, ren2026adapting, nie2025llada}. To generate a sequence $X = [x_1, \dots, x_n]$ of length $n$, a dLLM $f_\theta$ starts by a masked sequence $X_0 = [\texttt{mask}, \dots, \texttt{mask}]$. Then, it iteratively unmasks the tokens in the sequence until all are unmasked. Specifically, at each iteration $k$, $X_k$ is updated by forwarding $X_{k-1}$ through $f_\theta$, and replacing the masked tokens with the tokens that achieve logits higher than a threshold confidence hyperparameter $\gamma$, which can be multiple tokens at a time. If the logits at the masked tokens are not higher than $\gamma$, they remain for the next iteration. This inference pipeline achieves faster decoding speed than autoregressive LLM decoding~\cite{cai2026confidence, fu2026efficientdlm, liu2025sdlm, li2025prophet, mohamed2026sched}.


\textbf{Hypernetwork.}
A hypernetwork $h_\phi$ is a neural network trained to generate the parameters $\theta$ of a separate target network $f_\theta$~\cite{ha2017hypernetworks}. Formally, given an input or context $z$ (e.g., a task descriptor, layer index, or conditioning signal), the hypernetwork produces $\theta = h_\phi(z)$, and only $\phi$ is updated during training, while $\theta$ is instantiated on the fly for each $z$~\cite{ivison2022hyperdecoders}. Recent work attempt to shrink the hypernetwork's output to a low-dimensional parameter subspace rather than the full weight tensor~\cite{charakorn2025texttolora, charakorn2026doctolora}. In particular, hypernetworks have been used to generate low-rank adaptation (LoRA~\cite{hu2022lora}) updates $\Delta\theta = h_\phi(z) = A_z B_z^\top$, with $A_z \in \mathbb{R}^{d \times r}$, $B_z \in \mathbb{R}^{k \times r}$, $r \ll \min(d,k)$, reducing the hypernetwork's output dimensionality from $\mathcal{O}(dk)$ to $\mathcal{O}(r(d+k))$ per layer. More background can be found in the Appendix~\ref{appx:background}

\section{HyperZip's Framework}

In this section, we introduce \textsc{HyperZip}, a novel, efficient, and scalable framework for personalized LLM-based data compression.

\begin{wrapfigure}{r}{0.55\textwidth}
    \centering
    \includegraphics[width=\linewidth]{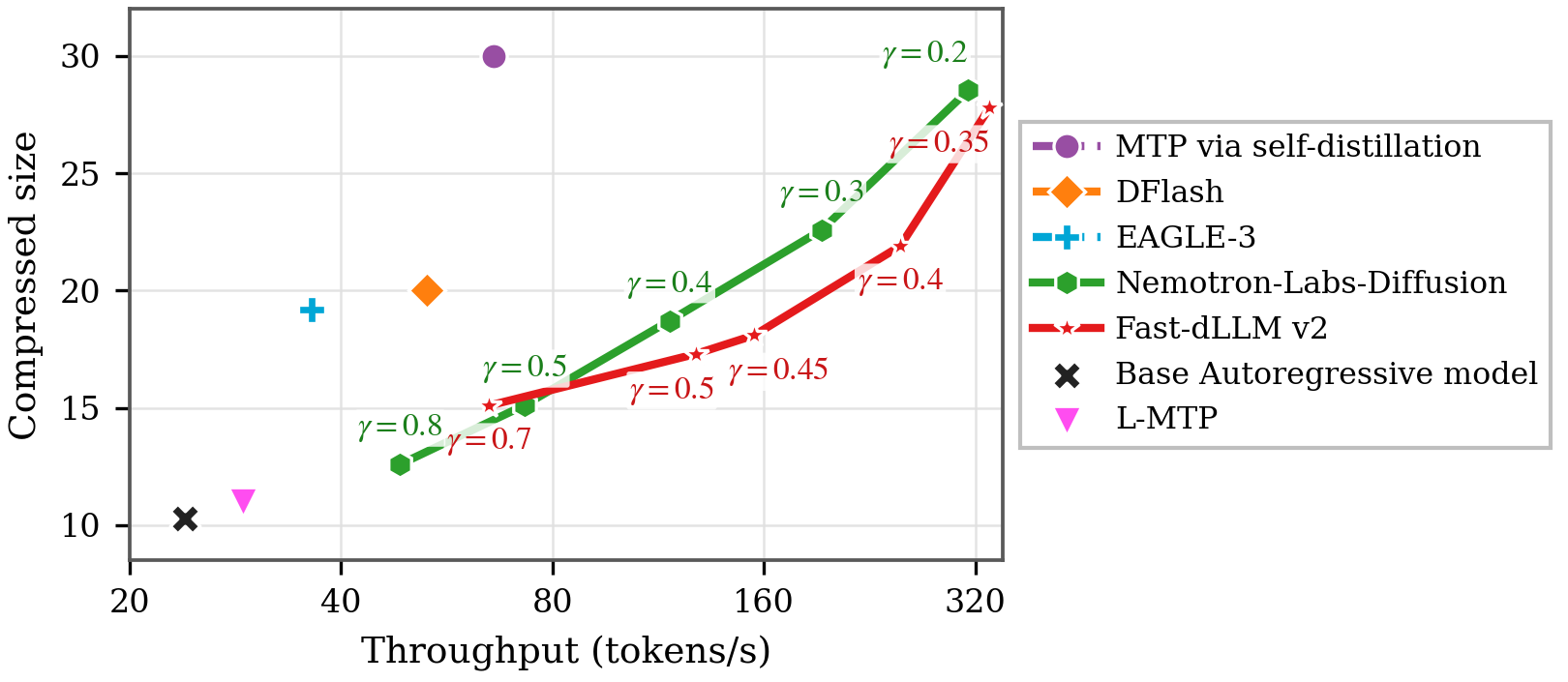}
    \caption{Compression--throughput trade-off across MTP mechanisms.}
    \label{fig:prelim-main}
\end{wrapfigure}

\textbf{Trade-off of compression rate and throughput.} We first present our preliminary results to characterize the trade-off between compression rate and speed incurred by MTP and dLLMs. We consider two main categories following~\cite{zhang2025parallel}: \textbf{(1)} AR-Based which includes \textbf{EAGLE-3}~\cite{li2025eagle3}, \textbf{L-MTP}~\cite{liu2025lmtp}, and \textbf{self-distillation}~\cite{kirchenbauer2026multitoken}; and \textbf{(2)} Non-AR-based which include \textbf{DFlash}~\cite{chen2026dflash}, \textbf{Nemotron-Labs-Diffusion}~\cite{fu2026nemotrondiffusion}, and \textbf{Fast-dLLM-v2}~\cite{wu2025fastdllmv2}. Figure~\ref{fig:prelim-main} illustrates the compression ratio and throughput of the considered mechanisms. In general, MTP mechanisms achieve higher throughput than autoregresisve generation. Specifically, by decreasing the confidence threshold $\gamma$, diffusion-based methods achieve the highest inference speed, but they incur a higher compression ratio. The key reason is sampler method choose to unmask sequence with lower confidence about the current context, so it requires a longer bitstring to compress the target data file. This behavior of dLLMs highlights a trade-off between compression rate and throughput, which can be addressed by aligning the dLLMs with the data, increasing its confidence for each token at the decoding process. Complete settings and results are provided in Appendix~\ref{app:mtp-study}.


\begin{wrapfigure}{r}{0.65\textwidth}
    \centering
    \includegraphics[width=\linewidth]{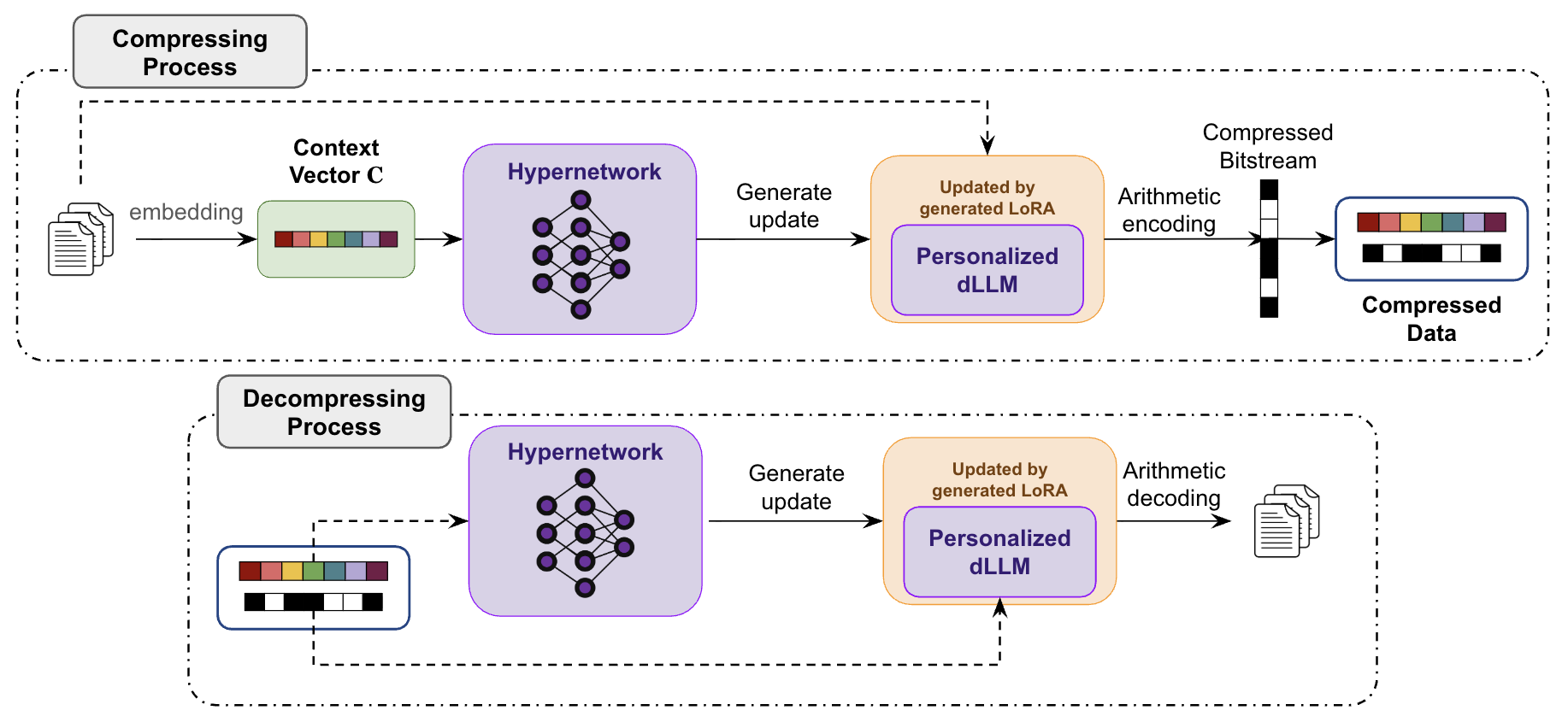}
    \caption{\textsc{HyperZip}'s pipeline.}
    \label{fig:setting}
\end{wrapfigure}

\textbf{Main pipeline.} To address the trade-off between compression ratio and throughput, \textsc{HyperZip} leverages a hypernetwork to generate updates for dLLMs at inference time to align them with the target data. Figure~\ref{fig:setting} illustrates this pipeline. To compress a target file containing a sequence $X \in V^n$, \textsc{HyperZip} uses an embedding model to extract a context vector $e = \texttt{emb}(X)$. Then, the hypernetwork $h_\phi$ takes $e$ as input to generate a LoRA update $\Delta_\theta$ for a dLLM $f_\theta$, personalizing it toward the target file. Using the updated dLLM, \textsc{HyperZip} incurs a multi-token prediction decoding process to get the logits of each token $x_i \in X, i\in [n]$, and encode it with arithmetic encoding into a bitstring representation. This bitstring, along with the context vector $e$, forms the compressed data. Unlike previous works~\cite{zhang2025l3tc}, \textsc{HyperZip} concatenates only the context vector with the bitstring, which is much smaller than an LLM's LoRA update, making the compressed data smaller and more efficient to transmit. To decompress, $e$ passes through the hypernetwork $h_\phi$ to update the dLLM $f_{\theta}$. Then, $f_{\theta}$ runs an arithmetic decoding process on a masked version of the target file, revealing each token if its logit exceeds $\gamma$. In addition, because dLLMs are personalized to the data, multiple tokens often have high confidence, so many tokens are revealed at each decoding step. This process can be parallelized, making decompression faster than conventional LLMs.

\textbf{Hypernetwork.} The core of \textsc{HyperZip} is a hypernetwork structure. Motivated by Text-to-LoRA~\cite{charakorn2025text}, \textsc{HyperZip} employs an embedding model $\texttt{emb}(\cdot)$ to extract a context vector $e$ from $X$, which serves as the main input for the hypernetwork $h_\phi$. In the training process, $\phi$ is learned given a training dataset $D$ by optimizing the following objective: $\arg\min_\phi\sum_{i=1}^{|D|}\ell\Big(x_i; \theta_0 + h_\phi(\texttt{emb}(x_i))\Big)$, where $\ell(\cdot;\cdot)$ is the loss function given the downstream task of $D$, and $\theta_0$ is the pretrained parameter of the dLLM. In addition, at inference time, the hypernetwork $\phi$ is fixed and can be deployed on a service provider server to generate updates to the dLLM for the compression and decompression processes.

\section{Experiments}

\textbf{Experimental Setting.} We conduct extensive experiments on the \textbf{FineWeb-Edu} dataset ~\cite{penedo2024fineweb} and \textbf{enwik9} dataset~\cite{hutter_enwik} to analyze the performance of \textsc{HyperZip}. We compare \textsc{HyperZip} against extensive baselines including the \textbf{Auto Regressive model}~\cite{yang2024qwen25}, \textbf{L3TC}~\cite{zhang2024l3tc}, \textbf{Fast-dLLM-v2}~\cite{wu2025fastdllmv2} and \textbf{standard supervised fine-tuning}. Appendix~\ref{app:exp-details} provides details on the baselines, evaluation metrics, and implementation.

\begin{wrapfigure}{r}{0.55\columnwidth}
    \centering
    {
        \includegraphics[
            width=\dimexpr0.75\linewidth-2\fboxrule-2\fboxsep\relax,
            trim={0 0 0 0},
            clip
        ]{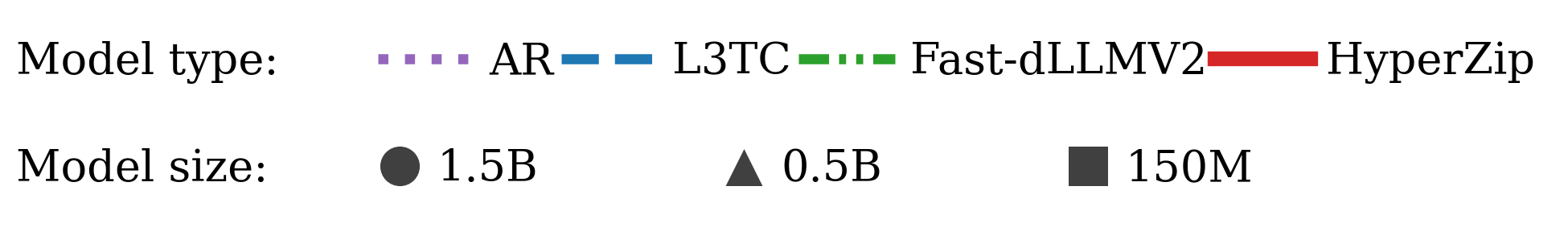}%
    }
    \begin{subfigure}[t]{0.49\linewidth}
        \centering
        \includegraphics[width=\linewidth]{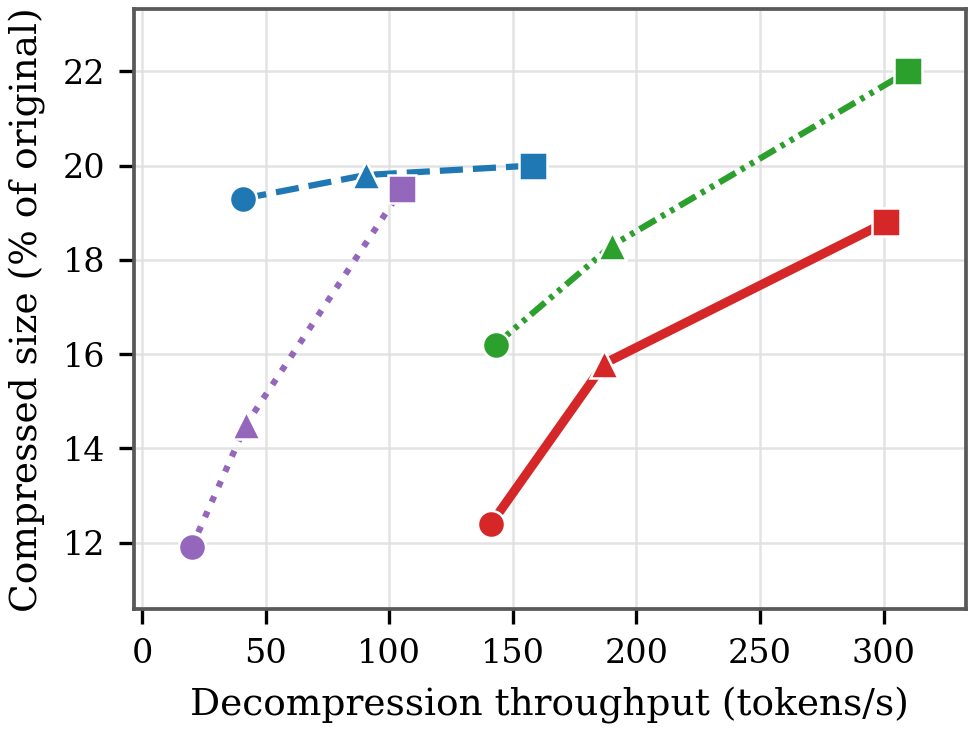}
        \caption{FineWeb-Edu}
        \label{fig:tradeoff-fineweb}
    \end{subfigure}
    \hfill
    \begin{subfigure}[t]{0.49\linewidth}
        \centering
        \includegraphics[width=\linewidth]{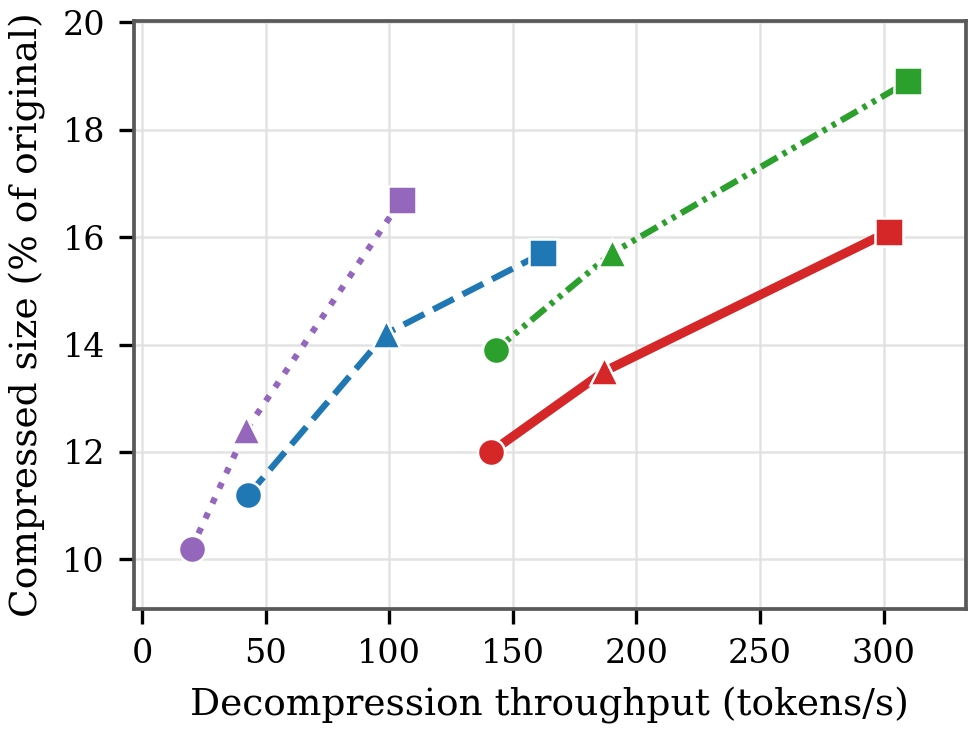}
        \caption{enwik9}
        \label{fig:tradeoff-enwiki9}
    \end{subfigure}

    \caption{Compression rate and throughput of \textsc{HyperZip} vs considered baselines.}
    \label{fig:compression-tradeoff}
\end{wrapfigure}


\textbf{Compression Ratio and Throughput.} In Figure~\ref{fig:compression-tradeoff}, we compare \textsc{HyperZip} with considered baselines in terms of compression ratio and decompression throughput. Lower compression ratio and higher throughput are better, so points closer to the bottom right indicate a better trade-off. We evaluate compression backbones ranging from 150M to 1.5B parameters and set the confidence threshold $\gamma=0.5$ for diffusion-based models. Overall, \textsc{HyperZip} achieves a better trade-off across the considered baselines by adapting the compression model to each. These results demonstrate \textsc{HyperZip} 's efficiency and its potential for real-world deployment.

\begin{wrapfigure}{r}{0.55\textwidth}
    \centering

    \captionof{table}{Compression ratio and throughput across document topics.}
    \label{tab:topic-compression-main}

    \footnotesize
    \setlength{\tabcolsep}{4pt}
    \renewcommand{\arraystretch}{1.08}

    \resizebox{\linewidth}{!}{%
    \begin{tabular}{@{}lccccc@{}}
        \toprule
        \multirow{2}{*}{\textbf{Model}}
        & \multicolumn{4}{c}{\textbf{CR incl. update vector} ($\downarrow$)}
        & \multirow{2}{*}{\begin{tabular}[c]{@{}c@{}}
            \textbf{Throughput}\\
            \textbf{(tok/s)} ($\uparrow$)
        \end{tabular}} \\
        \cmidrule(lr){2-5}
        & \textbf{Scientific}
          & \textbf{Literature}
          & \textbf{Programming}
          & \textbf{General Web}
          & \\
        \midrule

        Fast-dLLM
        & 16.1 & 15.7 & 18.0 & 16.3 & 141.0 \\

        Fine-tuning
        & 15.9 & 15.4 & 17.7 & 16.1 & 140.2 \\

        \midrule
        HyperZip (ours)
          & \textbf{15.6}
          & \textbf{15.3}
          & \textbf{16.3}
          & \textbf{15.8}
          & 138.9 \\

        \bottomrule
    \end{tabular}%
    }

    \vspace{0.8em}

    \includegraphics[
        width=0.98\linewidth,
        trim={0 0 0 0},
        clip
    ]{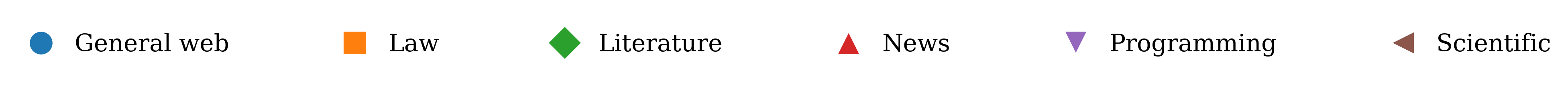}

    \vspace{0.2em}

    \begin{subfigure}[t]{0.49\linewidth}
        \centering
        \includegraphics[
            width=\linewidth
        ]{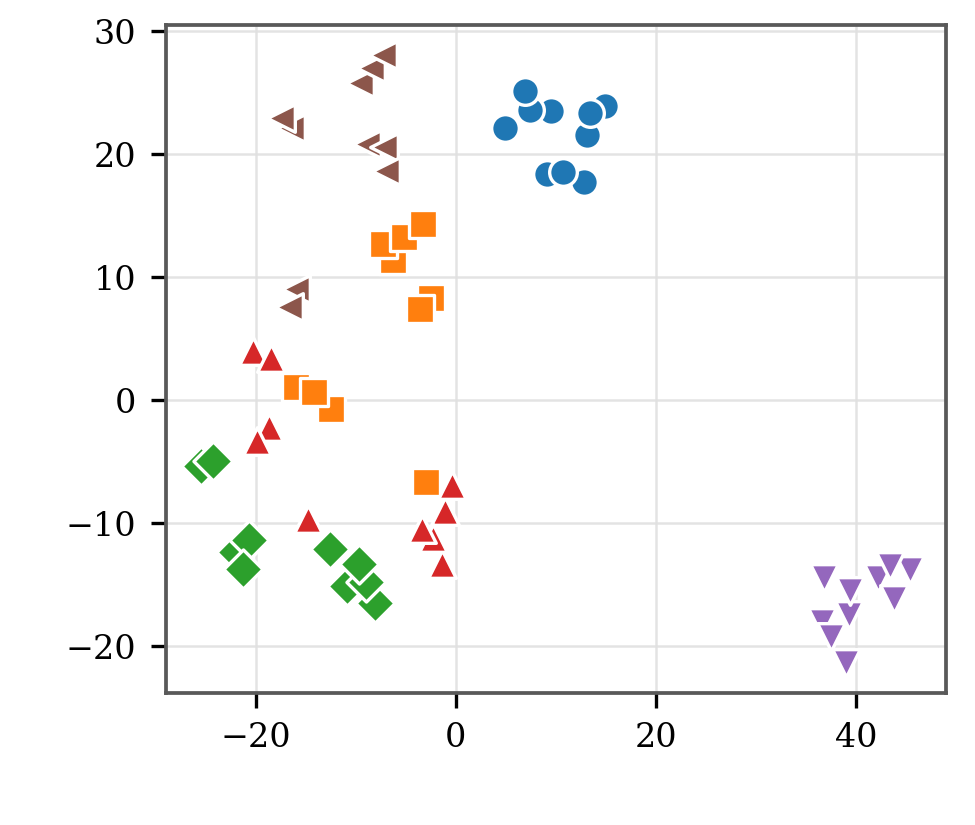}
        \caption{Context embedding}
        \label{fig:tsne-contextembed}
    \end{subfigure}
    \hfill
    \begin{subfigure}[t]{0.49\linewidth}
        \centering
        \includegraphics[
            width=\linewidth
        ]{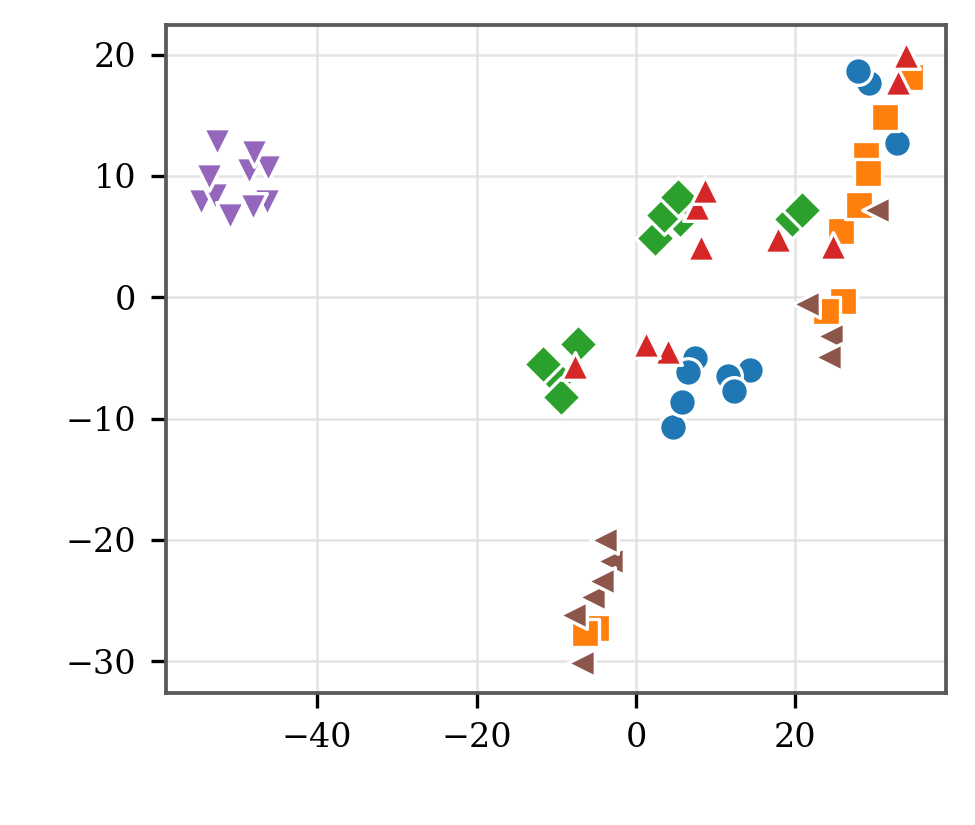}
        \caption{HyperZip output}
        \label{fig:tsne-hyperzip}
    \end{subfigure}

    \caption{t-SNE visualization.}
    \label{fig:generalize-visualization}
    \vspace{-10pt}
\end{wrapfigure}

\textbf{Generalizability.} We evaluate the generalization of \textsc{HyperZip} across different document topics, including law, scientific text, news, literature, programming, and general web text. Table~\ref{tab:topic-compression-main} shows that \textsc{HyperZip} consistently improves compression over the vanilla and fine-tuned dLLMs while maintaining similar throughput. For example, on programming text, it achieves a compression ratio of 16.3\% compared with 18.0\% for vanilla Fast-dLLM and 17.7\% for its fine-tuned version. We observe similar behavior for other topics. Its throughput remains comparable at 138.9 tok/s versus 141.0 and 140.2 tok/s. These results indicate \textsc{HyperZip} 's strong generalizability, highlighting its advantage in real-world scenarios where users have different data topics. In addition, Figure~\ref{fig:generalize-visualization} shows that the context vectors form clear topic-specific clusters, while \textsc{HyperZip} outputs exhibit greater overlap across topics, highlighting that \textsc{HyperZip} preserves topic-dependent information while mapping documents to a more shared representation. Therefore, \textsc{HyperZip} can be generalize for different topic as indicated by the results in Table~\ref{tab:topic-compression-main}.

\section{Conclusion}
 
In this work, we introduced \textsc{HyperZip}, an efficient and scalable framework that optimizes LLM-based lossless compression through diffusion models while preserving compression effectiveness through document-specific hypernetwork adaptation. Extensive experiments show that \textsc{HyperZip} achieves a better compression throughput trade-off than state-of-the-art baselines across datasets and model scales, making LLM-based compression more practical for real-world adoption. 

\bibliography{ref}

\bibliographystyle{abbrv}


\appendix


\section{Detailed Background and Preliminaries}\label{appx:background}
We consider streams of data $X = [x_1, x_2, \ldots, x_n] \in V^n$ of
length $n$ from a finite vocabulary $V$. The goal is compressing this data into the smallest reprentation. 

\subsection{LLMs in Data Compression}

\textbf{Coding Distribution.} Many lossless compression compressed based on coding distribution $\rho$. It is a sequence
of probability mass functions with $\rho_t : V^t \mapsto (0,1]$, $t \in
\mathbb{N}$, satisfying the chain rule
$
\rho_t(x_{1:t}) = \sum_{y \in V} \rho_{t+1}(x_{1:t}y)
$
for all $x_{1:t} \in V^t$, $t \le n$, and y is next token of the sequence. The conditional probability of a
symbol $x_t$ given past context $x_{1:t-1}$ is
$
\rho(x_t \mid x_{1:t-1}) := \rho(x_{1:t}) / \rho(x_{1:t-1}).
$

\textbf{Arithmetic Coding}
Arithmetic Coding~\cite{pasco1977source, rissanen1976generalized, witten1987arithmetic} encodes a sequence $x_{1:n}$ into a floating number using coding distribution $\rho$. The encoder narrows a nested sequence of sub-intervals of $[0,1)$, starting from $I_0 = [0,1)$. In step $k$, the current interval is partitioned into sub-intervals proportional to $\rho(\cdot \mid x_{1:k-1})$, and the one corresponding to the true symbol $x_k$ is retained as $I_k$ with correspond $\rho(x_k|x_{1:k-1})$. After $n$ steps, final interval $I_n$ uniquely identifies $x_{1:n}$, and its truncated binary expansion forms the encoded bitstream $c(x_{1:n})$. Its expected code length $L_\rho$ approaches the optimal Shannon’s bound: $L_\rho \geq H (\rho):=E_{x\sim\rho}[-log_2\rho(x)]$.~\cite{shannon1948mathematical}.

\textbf{LLM-Based Arithmetic Coding.}
Since the true source distribution $\rho$ is unknown, neural compression~\cite{yang2023introduction} approximates it with an autoregressive language model,
$
q_\theta(x_{1:n})=\prod_{i=1}^{n}q_\theta(x_i\mid x_{1:i-1})
$
~\cite{li2025lossless,mittu2024finezip,
heurteldepeiges2024compression,zhang2025testtime,
pan2025understanding}.
The arithmetic encoder uses $q_\theta(\cdot\mid x_{1:i-1})$, whose log-loss corresponds to the expected code length~\cite{deletang2024language}:
\[
\min_{\theta} L_{q_\theta}
=
\min_{\theta}
E_{x_{1:n}\sim \rho}
\left[
-\log_{2} q_\theta(x_{1:n})
\right]
=
\min_{\theta}
E_{x_{1:n}\sim \rho}
\left[
\sum_{i=1}^{n}
-\log_{2}q_\theta(x_i \mid x_{1:i-1})
\right].
\]

These methods outperform classical text compressors, including Gzip, LZMA, and Bzip2~\cite{zhang2024l3tc,valmeekam2023llmzip}, and can outperform PNG and FLAC on image and audio bytes~\cite{zhao2026omnizip,chen2025large}.

\subsection{Multi-Token Prediction with Diffusion Language Models}

\textbf{Basics of Diffusion Language Model.} Language Diffusion Models~\cite{austin2021structured,hoogeboom2021argmax, shi2024simplified, ren2026adapting, nie2025llada} are generative models produce text in parallel through an iterative denoising steps. Thus, compared to Auto Regressive generative model, DLMs shows efficientcy and advantage in reducing inference latency and capture bidirectional context. This field shows promising speed but burden with the generation capability.
Recent advances specificially block diffusion model~\cite{arriola2025block, fu2026nemotrondiffusion, wu2025fastdllmv2} set a new state-of-the-art performance among diffusion models on language modeling while preserve the speed up. People also introduce confidence-based sampler method. These new methods offer diffusion model a more stable generation while excel at latency optimization

\textbf{Inference process of DLMs and its MTP process.}
Divide the token sequence into $K$ ordered blocks,
\[
X = (X^{(1)}, \ldots, X^{(K)}).
\]
This mechanism processes blocks autoregressively while reconstructing
multiple tokens in parallel within each block. When processing block $k$
with sequence $X^{(k)} = (x_1^{(k)}, x_2^{(k)}, \ldots, x_{|X^{(k)}|}^{(k)})$,
the goal is to reconstruct the whole block using the minimum number of
steps. The preceding blocks $X^{(1:k-1)}$ are fixed, and the active block is
initialized as
\[
X_0^{(k)} = [\texttt{mask}]^{\otimes |X^{(k)}|}.
\]

At reconstruction step $s$, $X_s^{(k)}$ denotes the state of block $k$ at
step $s$. Define the remaining masked positions as
\[
\mathcal{M}_{k,s} = \big\{\, i : x_{s,i}^{(k)} = [\texttt{mask}] \,\big\}.
\]
For each $i \in \mathcal{M}_{k,s}$, the model produces a marginal distribution over
the vocabulary for that position, conditioned on the fixed prefix and the
current partially-denoised block:
\[
q_\theta(v)
=
q_\theta\!\left(x_{s,i}^{(k)} = v \;\middle|\; X^{(1:k-1)}, X_s^{(k)}\right),
\qquad v \in V.
\]
A shared deterministic rule then selects, among the still-masked
positions, which ones to commit at this step -- typically those where the
model's top prediction is confident enough:
\[
U_{k,s}
=
\big\{\, i \in \mathcal{M}_{k,s} :
\max_{v \in V} q_\theta(v) > \gamma \,\big\}.
\]
Positions in $U_{k,s}$ are unmasked (fixed to their argmax value), the
state is updated to $X_{s+1}^{(k)}$, and the process repeats until
$\mathcal{M}_{k,s} = \emptyset$.

\subsection{Hypernetwork in LLMs.}
A hypernetwork $h_\phi$ is a neural network trained to generate the
parameters of a separate target network $f_\theta$, rather than learning those
parameters directly via gradient descent on the target's own
loss~\cite{ha2017hypernetworks}. Formally, given an input or context $z$
(e.g., a task descriptor, layer index, or conditioning signal), the
hypernetwork produces $\Delta\theta = h_\phi(z)$, and only $\phi$ is
updated during training, while $\Delta\theta$ is instantiated on the fly
for each $z$~\cite{ivison2022hyperdecoders}. This decouples the number of
trainable parameters from the size of the target network, and allows the
target's effective parameters to vary dynamically with $z$ without
storing or fine-tuning a separate copy of $f_\theta$ for each context.

Applied naively to LLM-scale target networks, however, this formulation
requires $h_\phi$ to output on the order of $10^9$--$10^{11}$ parameters,
making direct weight generation computationally prohibitive and
empirically unstable at scale~\cite{dhankhar2026scaling}.
Recent work addresses this by restricting the hypernetwork's output to a
low-dimensional parameter subspace rather than the full weight
tensor~\cite{charakorn2025texttolora, charakorn2026doctolora}. In
particular, hypernetworks have been used to generate low-rank adaptation
(LoRA~\cite{hu2022lora}) updates $\Delta\theta = h_\phi(z) = A_z B_z^\top$,
with $A_z \in \mathbb{R}^{d \times r}$, $B_z \in \mathbb{R}^{k \times r}$,
$r \ll \min(d,k)$, reducing the hypernetwork's output dimensionality from
$\mathcal{O}(dk)$ to $\mathcal{O}(r(d+k))$ per layer. This enables input-
or task-conditioned weight generation at LLM scale, allowing $\Delta\theta$
to be produced in a single forward pass of $h_\phi$ rather than requiring
gradient-based fine-tuning for each new context
$z$.
\section{Detailed Multi-Token Prediction Study} \label{app:mtp-study} 



\subsection{Experimental Setup.} Whenever supported, we use Qwen3-4B~\cite{yang2025qwen3} as the common backbone and evaluate all methods on \textbf{enwiki8}~\cite{hutter_enwik} using the same settings with 1 GPU A100, precision, batch size, and sequence length. We report compression ratio, compression throughput in tokens per second, and the average number of tokens processed per forward pass. Unlike conventional speculative generation, where drafted tokens must be verified by a target model~\cite{leviathan2023fast,chen2023accelerating}, lossless compression directly provides the ground-truth sequence. Drafted tokens can therefore be compared with the observed tokens without a separate verification-model invocation. 
\begin{figure}[t]
    \centering
    \includegraphics[width=\linewidth]{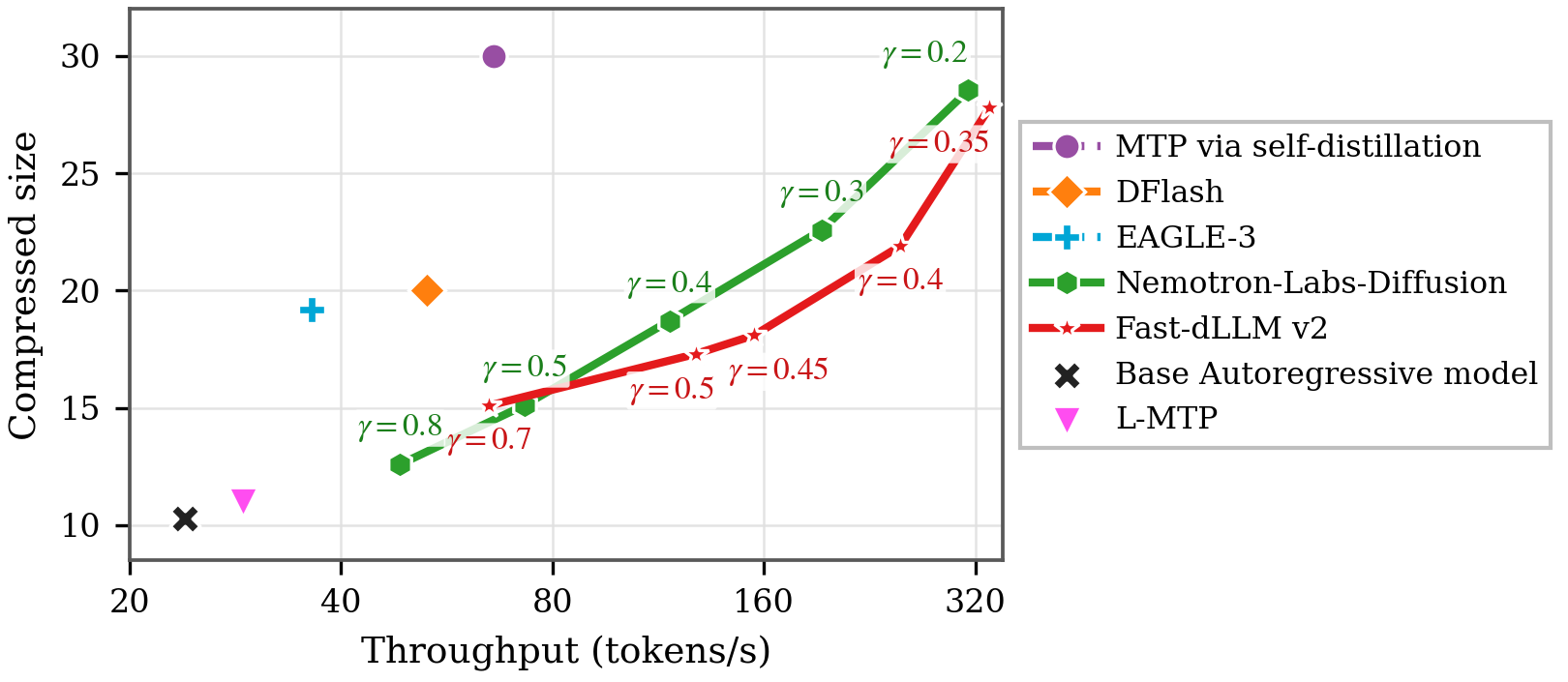}
    \caption{Compression--throughput trade-off across inference methods and
    configurations. Lower compressed size and higher throughput are preferred.
    Each curve connects configurations from the corresponding hyperparameter
    sweep. The throughput axis uses a base-2 logarithmic scale. Detail of how these methods custom speed present in Baseline}
    \label{fig:prelim}
\end{figure} 

\subsection{Baselines.}

We briefly describe each baseline and the hyperparameter that controls its decoding speed.

\textbf{MTP via Self-Distillation}~\cite{kirchenbauer2026multitoken}
predicts a block of future tokens in parallel. Increasing the block size allows more
tokens to be generated per forward pass, potentially improving throughput at the
cost of prediction accuracy. We evaluate block sizes of $\{1,10,15,20\}$.

\textbf{EAGLE-3}~\cite{li2025eagle3}
uses a lightweight draft model to propose multiple future tokens, which are then
verified in parallel by the target model. We control its decoding speed by varying
the number of drafted tokens per iteration; longer drafts provide more opportunities
for acceleration but may reduce the acceptance rate.

\textbf{DFlash}~\cite{chen2026dflash}
combines an autoregressive target model with a diffusion-based drafter that proposes
a block of future tokens in parallel. We use a five-layer draft model and vary the
block size over $\{5,10,15,20\}$ to obtain different throughput operating points.

\textbf{L-MTP}~\cite{liu2025lmtp}
augments an autoregressive model with multiple MLP prediction heads, where the number
of heads determines how many future tokens can be predicted in parallel. Because
different head configurations require separate training, L-MTP cannot flexibly adjust
its decoding speed at inference time. We therefore evaluate a single configuration
with four prediction heads, whose outputs are verified in parallel.

\textbf{Autoregressive Model}
generates one token at a time, conditioning each prediction on all preceding tokens.
Unlike parallel decoding methods, it has no hyperparameter for varying the number of
tokens generated per forward pass and therefore serves as our single-token decoding
baseline.

\textbf{Nemotron-Labs-Diffusion}~\cite{fu2026nemotrondiffusion}
generates tokens within each block in parallel through iterative denoising. We vary the confidence threshold over ${0.8, 0.5, 0.4, 0.3, 0.2}$ to obtain different decoding speeds. Lower thresholds allow more tokens to be accepted at each denoising step, increasing throughput but potentially reducing prediction quality and compression effectiveness.

\textbf{Fast-dLLM-v2}~\cite{wu2025fastdllmv2}
also uses confidence-based parallel decoding. We vary its confidence threshold over ${0.7, 0.5, 0.45, 0.4, 0.35}$ to evaluate different throughput operating points. As with Nemotron-Labs-Diffusion, lower thresholds increase decoding parallelism but may worsen compression effectiveness.

\subsection{Results and Analysis.}

Figure~\ref{fig:prelim} reveals a clear trade-off between decompression throughput and compression effectiveness. All parallel decoding methods improve throughput over the autoregressive baseline, but generally produce larger compressed representations. Overall, diffusion based methods, including Nemotron-Labs-Diffusion~\cite{fu2026nemotrondiffusion} and Fast-dLLM-v2~\cite{wu2025fastdllmv2}, provide a more favorable trade-off than AR-based methods, particularly at high-throughput operating points.

Among the AR-based methods, MTP via self-distillation increases throughput from approximately 25 to 180 tokens/s, while its compressed size rises from 12\% to 42\%. EAGLE-3 reaches approximately 50 tokens/s, but its compressed size increases sharply with draft length. These results suggest that aggressively predicting more future tokens with auxiliary prediction heads can substantially degrade the probability estimates used by the arithmetic coder. Although DFlash employs a diffusion-based drafter, it follows a similar trend: throughput increases from approximately 50 to 170 tokens/s, while compressed size grows from 20\% to 42\%. This suggests that adding a diffusion drafter can accelerate decoding but remains constrained by the underlying autoregressive generation process.

In contrast, diffusion-based methods maintain better compression effectiveness at
comparable throughput. Nemotron-Labs-Diffusion reaches approximately
140 tokens/s with a compressed size below 20\%, whereas Fast-dLLM-v2
achieves approximately 128 tokens/s with a compressed size of 17.3\%.
At more aggressive settings, Fast-dLLM-v2 further increases throughput to
approximately 320 tokens/s, although its compressed size rises to about
28\%. Thus, Fast-dLLM-v2 provides the strongest overall trade-off among the
evaluated methods rather than uniformly achieving the best compression at
every operating point.

We hypothesize that this advantage arises from the flexibility of confidence-based diffusion decoding. Its throughput can be adjusted at inference time by changing the confidence threshold, whereas AR-based methods are more strongly constrained by the prediction horizon and drafting configuration used during training~\cite{xiang2026pmtp}.
\section{Experiment Details} \label{app:exp-details}
\subsection{Model Architecture}
\label{app:model-architecture}

\paragraph{Diffusion language model.}
Our compression backbones follow the architectures and pretrained weights of
BD3LM and Fast-dLLM-v2 where they will distill pretrained autoregressive model to get the diffusion language model~\cite{arriola2025block,wu2025fastdllmv2}. We evaluate
three model scales: 150M, 0.5B, and 1.5B parameters distill from SmolLM2-135M~\cite{allal2025smollm2}. The backbone parameters
remain frozen during hypernetwork training.

Each backbone operates on blocks of tokens and iteratively unmasks tokens
according to the confidence of its predictions. Its decoding throughput is
controlled primarily by two hyperparameters: the block size \(B\) and the
confidence threshold $\gamma$. A larger \(B\) provides more opportunities for
parallel prediction, while a lower $\gamma$ permits more tokens to be unmasked
during each model evaluation. Both changes can improve throughput, but may
produce less confident probability estimates and therefore increase the
resulting arithmetic-coded bitstream.

The original backbones are trained with a block size of \(32\). We adapt them
to a block size of \(256\) using the training procedure described in
Appendix~\ref{app:training-details}, thereby enabling greater decoding
parallelism. During training, we set $\gamma=1$, which disables confidence-based
parallel unmasking and ensures that the training objective is not affected by
the inference-time decoding policy. Unless otherwise stated, we use
$\gamma =0.5$ at inference time, as it provides the best empirical trade-off
between compression effectiveness and throughput on the validation set. 

Table~\ref{tab:backbone-configurations} summarizes the backbone configurations.

\begin{table}[t]
    \centering
    \small
    \caption{Diffusion-language-model configurations used in our experiments.
    Replace the placeholders with the exact checkpoint information.}
    \label{tab:backbone-configurations}
    \begin{tabular}{lcccc}
        \toprule
        Backbone & Parameters & Layers & Hidden size & Vocabulary \\
        \midrule
        \texttt{BD3LM-135M} & 135M & \texttt{30} & \texttt{576} & \texttt{49152} \\
        \texttt{BD3LM-0.5B} & 0.5B & \texttt{24} & \texttt{896} & \texttt{151936} \\
        \texttt{Fast-dLLM-V2-1.5B} & 1.5B & \texttt{28} & \texttt{1536} & \texttt{151936} \\
        \bottomrule
    \end{tabular}
\end{table}

\paragraph{Hypernetwork.}
Our hypernetwork is inspired by Text-to-LoRA~\cite{charakorn2025text} and
contains two components: (i) a document encoder that produces a semantic
representation of the input document and (ii) a multilayer perceptron (MLP)
that generates document-specific LoRA parameters for the frozen diffusion
backbone.

For each document \(d\), we obtain a document representation
\(\mathbf{e}_{d}\in\mathbb{R}^{1536}\) using
GTE-Qwen2-1.5B-Instruct\footnote{\url{https://huggingface.co/Alibaba-NLP/gte-Qwen2-1.5B-instruct}}.
For documents longer than the encoder's maximum input length, we truncate the document to the first 32,000 tokens / divide it into
chunks.

To distinguish the LoRA parameters generated for different locations in the
backbone, we associate each adapted module \(m\) and layer \(\ell\) with
learned embeddings
\(\mathbf{e}_{m}\in\mathbb{R}^{32}\) and
\(\mathbf{e}_{\ell}\in\mathbb{R}^{32}\), respectively. These embeddings
are concatenated with the document representation:
\[
    z
    =
    [\mathbf{e}_{d};
     \mathbf{e}_{m};
     \mathbf{e}_{\ell}].
\]
The resulting conditioning vector is passed through an \(L\)-layer MLP:
\[
    \Delta\theta
    =
    h_{\phi}(\mathbf{z}),
\]
where \(h_{\phi}\) denotes the shared hypernetwork.

For a target weight matrix
\(\mathbf{W}\in\mathbb{R}^{d_{\mathrm{out}}\times
d_{\mathrm{in}}}\), the hypernetwork produces the two LoRA factors
\[
    \mathbf{A}
        \in\mathbb{R}^{r\times d_{\mathrm{in}}},
    \qquad
    \mathbf{B}
        \in\mathbb{R}^{d_{\mathrm{out}}\times r}.
\]
The personalized weight is then
\[
    \mathbf{W}'
    =
    \mathbf{W}
    +
    \frac{\alpha}{r}
    \mathbf{B}\mathbf{A},
\]
where \(r\) is the LoRA rank and \(\alpha\) is the LoRA scaling factor.
We generate adaption of each
transformer layer. All backbone weights remain frozen; only the hypernetwork,
layer embeddings, and module embeddings are optimized.

Table~\ref{tab:hypernetwork-configurations} reports the complete
hypernetwork configuration.

\begin{table}[t]
    \centering
    \small
    \caption{Hypernetwork configuration.}
    \label{tab:hypernetwork-configurations}
    \begin{tabular}{lc}
        \toprule
        Component & Configuration \\
        \midrule
        Document encoder & GTE-Qwen2-1.5B-Instruct \\
        Document-embedding dimension & 1536 \\
        Encoder input length & \texttt{32,000} tokens \\
        Module-embedding dimension \(d_m\) & \texttt{32} \\
        Layer-embedding dimension \(d_\ell\) & \texttt{32} \\
        LoRA rank \(r\) & \texttt{8} \\
        Target modules & \texttt{\(q\), \(v\) projections} \\
        Hypernetwork parameter & \texttt{55M} \\
        \bottomrule
    \end{tabular}
\end{table}

\subsection{Training Details}
\label{app:training-details}

\paragraph{Meta-training data.}
We construct the meta-training corpus from a subset of
FineWeb-Edu~\cite{penedo2024fineweb}, treating each sample as an individual
document \(d\). The resulting training corpus contains approximately
900 million tokens after tokenization with the backbone tokenizer. We construct
a disjoint FineWeb-Edu evaluation split containing 100,000 documents and
approximately 120 million tokens. We remove any document appearing in the
training split and additionally perform deduplication to reduce potential
overlap between the training and evaluation sets.

We also evaluate on enwik9~\cite{hutter_enwik} with no enwik9 text is used for
hypernetwork training, hyperparameter selection, or early stopping.

\paragraph{Training objective.}
For every document \(c\), the document encoder first produces
\(\mathbf{e}_{c}\). The hypernetwork then generates the LoRA factors used to
personalize the frozen diffusion backbone. The personalized model is trained
with the same masked diffusion objective as the underlying backbone:
\[
    \mathcal{L}(\theta)
    =
    \mathbb{E}_{d,t,\widetilde{\mathbf{x}}_t}
    \left[
        -\sum_{i\in\mathcal{M}_t}
        \log q_{\theta}
        \left(
            x_i
            \mid
            \widetilde{\mathbf{x}}_t,
            \mathbf{e}_{d}
        \right)
    \right],
\]
where \(\mathcal{M}_t\) is the set of masked positions at diffusion step \(t\),
\(\widetilde{\mathbf{x}}_t\) is the corrupted sequence, and the backbone
parameters are fixed. Gradients are propagated only through the generated
LoRA parameters and the trainable hypernetwork components.

\paragraph{Optimization.}
We train for 3,500 optimization steps with a global batch size of 128 and a
learning rate of \(2\times10^{-5}\). Training is performed on four NVIDIA A100
GPUs using bfloat16 precision. The learning
rate follows a cosine schedule with
warm-up fraction of 0.1.

We use sequences of length 8192, a block size of \(256\), and a
sub-block size of \(8\). The confidence threshold is fixed at $\gamma=1$
during training.

\subsection{Compression and Evaluation Protocol}
\label{app:evaluation-protocol}

\paragraph{Lossless compression procedure.}
We tokenize each document using the tokenizer associated with its compression
backbone. The personalized diffusion model produces the conditional
probabilities used by the arithmetic coder. Encoding and decoding use
identical model parameters, tokenization, adapter-generation procedures, and
probability quantization to guarantee deterministic reconstruction. We verify
losslessness by checking that the decoded byte sequence exactly matches the
original input.

\subsection{Evaluation Metrics}
\label{app:metrics}

We evaluate the trade-off between compression effectiveness and processing
speed. Lower compression ratios and higher throughput are preferred.

\paragraph{Compression ratio.}
Let \(S_{\mathrm{raw}}\) denote the size of the original document in bytes and
\(S_{\mathrm{payload}}\) the size of its arithmetic-coded representation. We
report the payload compression ratio as
\begin{equation}
    \mathrm{CR}_{\mathrm{payload}}
    =
    100\times
    \frac{S_{\mathrm{payload}}}{S_{\mathrm{raw}}}.
\end{equation}
This metric is expressed as a percentage; a lower value indicates better
compression. For example, \(\mathrm{CR}=12\%\) means that the compressed
payload occupies 12\% of the original file size.



\paragraph{Compression and decompression throughput.}
Throughput is measured in tokens per second:
\begin{equation}
    \mathrm{Throughput}
    =
    \frac{N_{\mathrm{tokens}}}{T},
\end{equation}



\begin{table*}[t]
    \centering
    \caption{Compression ratio across document topics and compression
    throughput. Lower compression ratio and higher throughput are preferred.}
    \label{tab:topic-compression}
    \footnotesize
    \setlength{\tabcolsep}{4pt}
    \renewcommand{\arraystretch}{1.08}

    \resizebox{0.7\textwidth}{!}{%
    \begin{tabular}{@{}lcccccc@{}}
        \toprule
        \multirow{2}{*}{\textbf{Model}}
        & \multirow{2}{*}{\boldmath$\gamma$}
        & \multicolumn{4}{c}{\textbf{CR incl. update vector} ($\downarrow$)}
        & \multirow{2}{*}{\begin{tabular}[c]{@{}c@{}}
            \textbf{Throughput}\\
            \textbf{(tok/s)} ($\uparrow$)
        \end{tabular}} \\
        \cmidrule(lr){3-6}
        & & \textbf{Scientific}
          & \textbf{Literature}
          & \textbf{Programming}
          & \textbf{General Web}
          & \\
        \midrule

        Fast-dLLM
          & 0.5 & 16.1 & 15.7 & 18.0 & 16.3 & 141.0 \\

        FineZip
          & -- & 9.23 & 8.46 & 9.24 & 8.22 & 12.6 \\

        Fine-tuning
          & 0.5 & 15.9 & 15.4 & 17.7 & 16.1 & 140.2 \\

        \midrule
        HyperZip (ours)
          & 0.5
          & \textbf{15.6}
          & \textbf{15.3}
          & \textbf{16.3}
          & \textbf{15.8}
          & \textbf{138.9} \\

        \bottomrule
    \end{tabular}%
    }
\end{table*}


\newpage
\input{checklist.tex}

\end{document}

%% file: checklist.tex
\section*{NeurIPS Paper Checklist}

The checklist is designed to encourage best practices for responsible machine learning research, addressing issues of reproducibility, transparency, research ethics, and societal impact. Do not remove the checklist: {\bf The papers not including the checklist will be desk rejected.} The checklist should follow the references and follow the (optional) supplemental material.  The checklist does NOT count towards the page
limit. 

Please read the checklist guidelines carefully for information on how to answer these questions. For each question in the checklist:
\begin{itemize}
    \item You should answer \answerYes{}, \answerNo{}, or \answerNA{}.
    \item \answerNA{} means either that the question is Not Applicable for that particular paper or the relevant information is Not Available.
    \item Please provide a short (1--2 sentence) justification right after your answer (even for \answerNA). 
\end{itemize}

{\bf The checklist answers are an integral part of your paper submission.} They are visible to the reviewers, area chairs, senior area chairs, and ethics reviewers. You will also be asked to include it (after eventual revisions) with the final version of your paper, and its final version will be published with the paper.

The reviewers of your paper will be asked to use the checklist as one of the factors in their evaluation. While \answerYes{} is generally preferable to \answerNo{}, it is perfectly acceptable to answer \answerNo{} provided a proper justification is given (e.g., error bars are not reported because it would be too computationally expensive'' or ``we were unable to find the license for the dataset we used''). In general, answering \answerNo{} or \answerNA{} is not grounds for rejection. While the questions are phrased in a binary way, we acknowledge that the true answer is often more nuanced, so please just use your best judgment and write a justification to elaborate. All supporting evidence can appear either in the main paper or the supplemental material, provided in appendix. If you answer \answerYes{} to a question, in the justification please point to the section(s) where related material for the question can be found.

IMPORTANT, please:
\begin{itemize}
    \item {\bf Delete this instruction block, but keep the section heading ``NeurIPS Paper Checklist"},
    \item  {\bf Keep the checklist subsection headings, questions/answers and guidelines below.}
    \item {\bf Do not modify the questions and only use the provided macros for your answers}.
\end{itemize}


\begin{enumerate}

\item {\bf Claims}
    \item[] Question: Do the main claims made in the abstract and introduction accurately reflect the paper's contributions and scope?
    \item[] Answer: \answerYes{} 
    \item[] Guidelines:
    \begin{itemize}
        \item The answer \answerNA{} means that the abstract and introduction do not include the claims made in the paper.
        \item The abstract and/or introduction should clearly state the claims made, including the contributions made in the paper and important assumptions and limitations. A \answerNo{} or \answerNA{} answer to this question will not be perceived well by the reviewers. 
        \item The claims made should match theoretical and experimental results, and reflect how much the results can be expected to generalize to other settings. 
        \item It is fine to include aspirational goals as motivation as long as it is clear that these goals are not attained by the paper. 
    \end{itemize}

\item {\bf Limitations}
    \item[] Question: Does the paper discuss the limitations of the work performed by the authors?
    \item[] Answer: \answerNo{} 
    \item[] Guidelines:
    \begin{itemize}
        \item The answer \answerNA{} means that the paper has no limitation while the answer \answerNo{} means that the paper has limitations, but those are not discussed in the paper. 
        \item The authors are encouraged to create a separate ``Limitations'' section in their paper.
        \item The paper should point out any strong assumptions and how robust the results are to violations of these assumptions (e.g., independence assumptions, noiseless settings, model well-specification, asymptotic approximations only holding locally). The authors should reflect on how these assumptions might be violated in practice and what the implications would be.
        \item The authors should reflect on the scope of the claims made, e.g., if the approach was only tested on a few datasets or with a few runs. In general, empirical results often depend on implicit assumptions, which should be articulated.
        \item The authors should reflect on the factors that influence the performance of the approach. For example, a facial recognition algorithm may perform poorly when image resolution is low or images are taken in low lighting. Or a speech-to-text system might not be used reliably to provide closed captions for online lectures because it fails to handle technical jargon.
        \item The authors should discuss the computational efficiency of the proposed algorithms and how they scale with dataset size.
        \item If applicable, the authors should discuss possible limitations of their approach to address problems of privacy and fairness.
        \item While the authors might fear that complete honesty about limitations might be used by reviewers as grounds for rejection, a worse outcome might be that reviewers discover limitations that aren't acknowledged in the paper. The authors should use their best judgment and recognize that individual actions in favor of transparency play an important role in developing norms that preserve the integrity of the community. Reviewers will be specifically instructed to not penalize honesty concerning limitations.
    \end{itemize}

\item {\bf Theory assumptions and proofs}
    \item[] Question: For each theoretical result, does the paper provide the full set of assumptions and a complete (and correct) proof?
    \item[] Answer: \answerNA{} 
    \item[] Justification: We do not do theoretical analysis. 
    \item[] Guidelines:
    \begin{itemize}
        \item The answer \answerNA{} means that the paper does not include theoretical results. 
        \item All the theorems, formulas, and proofs in the paper should be numbered and cross-referenced.
        \item All assumptions should be clearly stated or referenced in the statement of any theorems.
        \item The proofs can either appear in the main paper or the supplemental material, but if they appear in the supplemental material, the authors are encouraged to provide a short proof sketch to provide intuition. 
        \item Inversely, any informal proof provided in the core of the paper should be complemented by formal proofs provided in appendix or supplemental material.
        \item Theorems and Lemmas that the proof relies upon should be properly referenced. 
    \end{itemize}

    \item {\bf Experimental result reproducibility}
    \item[] Question: Does the paper fully disclose all the information needed to reproduce the main experimental results of the paper to the extent that it affects the main claims and/or conclusions of the paper (regardless of whether the code and data are provided or not)?
    \item[] Answer: \answerYes{} 
    \item[] Guidelines:
    \begin{itemize}
        \item The answer \answerNA{} means that the paper does not include experiments.
        \item If the paper includes experiments, a \answerNo{} answer to this question will not be perceived well by the reviewers: Making the paper reproducible is important, regardless of whether the code and data are provided or not.
        \item If the contribution is a dataset and\slash or model, the authors should describe the steps taken to make their results reproducible or verifiable. 
        \item Depending on the contribution, reproducibility can be accomplished in various ways. For example, if the contribution is a novel architecture, describing the architecture fully might suffice, or if the contribution is a specific model and empirical evaluation, it may be necessary to either make it possible for others to replicate the model with the same dataset, or provide access to the model. In general. releasing code and data is often one good way to accomplish this, but reproducibility can also be provided via detailed instructions for how to replicate the results, access to a hosted model (e.g., in the case of a large language model), releasing of a model checkpoint, or other means that are appropriate to the research performed.
        \item While NeurIPS does not require releasing code, the conference does require all submissions to provide some reasonable avenue for reproducibility, which may depend on the nature of the contribution. For example
        \begin{enumerate}
            \item If the contribution is primarily a new algorithm, the paper should make it clear how to reproduce that algorithm.
            \item If the contribution is primarily a new model architecture, the paper should describe the architecture clearly and fully.
            \item If the contribution is a new model (e.g., a large language model), then there should either be a way to access this model for reproducing the results or a way to reproduce the model (e.g., with an open-source dataset or instructions for how to construct the dataset).
            \item We recognize that reproducibility may be tricky in some cases, in which case authors are welcome to describe the particular way they provide for reproducibility. In the case of closed-source models, it may be that access to the model is limited in some way (e.g., to registered users), but it should be possible for other researchers to have some path to reproducing or verifying the results.
        \end{enumerate}
    \end{itemize}

\item {\bf Open access to data and code}
    \item[] Question: Does the paper provide open access to the data and code, with sufficient instructions to faithfully reproduce the main experimental results, as described in supplemental material?
    \item[] Answer: \answerTODO{} 
    \item[] Justification: \justificationTODO{}
    \item[] Guidelines:
    \begin{itemize}
        \item The answer \answerNA{} means that paper does not include experiments requiring code.
        \item Please see the NeurIPS code and data submission guidelines (\url{https://neurips.cc/public/guides/CodeSubmissionPolicy}) for more details.
        \item While we encourage the release of code and data, we understand that this might not be possible, so \answerNo{} is an acceptable answer. Papers cannot be rejected simply for not including code, unless this is central to the contribution (e.g., for a new open-source benchmark).
        \item The instructions should contain the exact command and environment needed to run to reproduce the results. See the NeurIPS code and data submission guidelines (\url{https://neurips.cc/public/guides/CodeSubmissionPolicy}) for more details.
        \item The authors should provide instructions on data access and preparation, including how to access the raw data, preprocessed data, intermediate data, and generated data, etc.
        \item The authors should provide scripts to reproduce all experimental results for the new proposed method and baselines. If only a subset of experiments are reproducible, they should state which ones are omitted from the script and why.
        \item At submission time, to preserve anonymity, the authors should release anonymized versions (if applicable).
        \item Providing as much information as possible in supplemental material (appended to the paper) is recommended, but including URLs to data and code is permitted.
    \end{itemize}

\item {\bf Experimental setting/details}
    \item[] Question: Does the paper specify all the training and test details (e.g., data splits, hyperparameters, how they were chosen, type of optimizer) necessary to understand the results?
    \item[] Answer: \answerYes{} 
    \item[] Guidelines:
    \begin{itemize}
        \item The answer \answerNA{} means that the paper does not include experiments.
        \item The experimental setting should be presented in the core of the paper to a level of detail that is necessary to appreciate the results and make sense of them.
        \item The full details can be provided either with the code, in appendix, or as supplemental material.
    \end{itemize}

\item {\bf Experiment statistical significance}
    \item[] Question: Does the paper report error bars suitably and correctly defined or other appropriate information about the statistical significance of the experiments?
    \item[] Answer: \answerNo{} 
    \item[] Guidelines:
    \begin{itemize}
        \item The answer \answerNA{} means that the paper does not include experiments.
        \item The authors should answer \answerYes{} if the results are accompanied by error bars, confidence intervals, or statistical significance tests, at least for the experiments that support the main claims of the paper.
        \item The factors of variability that the error bars are capturing should be clearly stated (for example, train/test split, initialization, random drawing of some parameter, or overall run with given experimental conditions).
        \item The method for calculating the error bars should be explained (closed form formula, call to a library function, bootstrap, etc.)
        \item The assumptions made should be given (e.g., Normally distributed errors).
        \item It should be clear whether the error bar is the standard deviation or the standard error of the mean.
        \item It is OK to report 1-sigma error bars, but one should state it. The authors should preferably report a 2-sigma error bar than state that they have a 96\% CI, if the hypothesis of Normality of errors is not verified.
        \item For asymmetric distributions, the authors should be careful not to show in tables or figures symmetric error bars that would yield results that are out of range (e.g., negative error rates).
        \item If error bars are reported in tables or plots, the authors should explain in the text how they were calculated and reference the corresponding figures or tables in the text.
    \end{itemize}

\item {\bf Experiments compute resources}
    \item[] Question: For each experiment, does the paper provide sufficient information on the computer resources (type of compute workers, memory, time of execution) needed to reproduce the experiments?
    \item[] Answer: \answerYes{} 
    \item[] Guidelines:
    \begin{itemize}
        \item The answer \answerNA{} means that the paper does not include experiments.
        \item The paper should indicate the type of compute workers CPU or GPU, internal cluster, or cloud provider, including relevant memory and storage.
        \item The paper should provide the amount of compute required for each of the individual experimental runs as well as estimate the total compute. 
        \item The paper should disclose whether the full research project required more compute than the experiments reported in the paper (e.g., preliminary or failed experiments that didn't make it into the paper). 
    \end{itemize}
    
\item {\bf Code of ethics}
    \item[] Question: Does the research conducted in the paper conform, in every respect, with the NeurIPS Code of Ethics \url{https://neurips.cc/public/EthicsGuidelines}?
    \item[] Answer: \answerYes{} 
    \item[] Guidelines:
    \begin{itemize}
        \item The answer \answerNA{} means that the authors have not reviewed the NeurIPS Code of Ethics.
        \item If the authors answer \answerNo, they should explain the special circumstances that require a deviation from the Code of Ethics.
        \item The authors should make sure to preserve anonymity (e.g., if there is a special consideration due to laws or regulations in their jurisdiction).
    \end{itemize}

\item {\bf Broader impacts}
    \item[] Question: Does the paper discuss both potential positive societal impacts and negative societal impacts of the work performed?
    \item[] Answer: \answerYes{} 
    \item[] Guidelines:
    \begin{itemize}
        \item The answer \answerNA{} means that there is no societal impact of the work performed.
        \item If the authors answer \answerNA{} or \answerNo, they should explain why their work has no societal impact or why the paper does not address societal impact.
        \item Examples of negative societal impacts include potential malicious or unintended uses (e.g., disinformation, generating fake profiles, surveillance), fairness considerations (e.g., deployment of technologies that could make decisions that unfairly impact specific groups), privacy considerations, and security considerations.
        \item The conference expects that many papers will be foundational research and not tied to particular applications, let alone deployments. However, if there is a direct path to any negative applications, the authors should point it out. For example, it is legitimate to point out that an improvement in the quality of generative models could be used to generate Deepfakes for disinformation. On the other hand, it is not needed to point out that a generic algorithm for optimizing neural networks could enable people to train models that generate Deepfakes faster.
        \item The authors should consider possible harms that could arise when the technology is being used as intended and functioning correctly, harms that could arise when the technology is being used as intended but gives incorrect results, and harms following from (intentional or unintentional) misuse of the technology.
        \item If there are negative societal impacts, the authors could also discuss possible mitigation strategies (e.g., gated release of models, providing defenses in addition to attacks, mechanisms for monitoring misuse, mechanisms to monitor how a system learns from feedback over time, improving the efficiency and accessibility of ML).
    \end{itemize}
    
\item {\bf Safeguards}
    \item[] Question: Does the paper describe safeguards that have been put in place for responsible release of data or models that have a high risk for misuse (e.g., pre-trained language models, image generators, or scraped datasets)?
    \item[] Answer: \answerNA{} 
    \item[] Guidelines:
    \begin{itemize}
        \item The answer \answerNA{} means that the paper poses no such risks.
        \item Released models that have a high risk for misuse or dual-use should be released with necessary safeguards to allow for controlled use of the model, for example by requiring that users adhere to usage guidelines or restrictions to access the model or implementing safety filters. 
        \item Datasets that have been scraped from the Internet could pose safety risks. The authors should describe how they avoided releasing unsafe images.
        \item We recognize that providing effective safeguards is challenging, and many papers do not require this, but we encourage authors to take this into account and make a best faith effort.
    \end{itemize}

\item {\bf Licenses for existing assets}
    \item[] Question: Are the creators or original owners of assets (e.g., code, data, models), used in the paper, properly credited and are the license and terms of use explicitly mentioned and properly respected?
    \item[] Answer: \answerTODO{} 
    \item[] Justification: \justificationTODO{}
    \item[] Guidelines:
    \begin{itemize}
        \item The answer \answerNA{} means that the paper does not use existing assets.
        \item The authors should cite the original paper that produced the code package or dataset.
        \item The authors should state which version of the asset is used and, if possible, include a URL.
        \item The name of the license (e.g., CC-BY 4.0) should be included for each asset.
        \item For scraped data from a particular source (e.g., website), the copyright and terms of service of that source should be provided.
        \item If assets are released, the license, copyright information, and terms of use in the package should be provided. For popular datasets, \url{paperswithcode.com/datasets} has curated licenses for some datasets. Their licensing guide can help determine the license of a dataset.
        \item For existing datasets that are re-packaged, both the original license and the license of the derived asset (if it has changed) should be provided.
        \item If this information is not available online, the authors are encouraged to reach out to the asset's creators.
    \end{itemize}

\item {\bf New assets}
    \item[] Question: Are new assets introduced in the paper well documented and is the documentation provided alongside the assets?
    \item[] Answer: \answerYes{} 
    \item[] Justification: we will release the code when paper get accepted. 
    \item[] Guidelines:
    \begin{itemize}
        \item The answer \answerNA{} means that the paper does not release new assets.
        \item Researchers should communicate the details of the dataset\slash code\slash model as part of their submissions via structured templates. This includes details about training, license, limitations, etc. 
        \item The paper should discuss whether and how consent was obtained from people whose asset is used.
        \item At submission time, remember to anonymize your assets (if applicable). You can either create an anonymized URL or include an anonymized zip file.
    \end{itemize}

\item {\bf Crowdsourcing and research with human subjects}
    \item[] Question: For crowdsourcing experiments and research with human subjects, does the paper include the full text of instructions given to participants and screenshots, if applicable, as well as details about compensation (if any)? 
    \item[] Answer: \answerNA{} 
    \item[] Guidelines:
    \begin{itemize}
        \item The answer \answerNA{} means that the paper does not involve crowdsourcing nor research with human subjects.
        \item Including this information in the supplemental material is fine, but if the main contribution of the paper involves human subjects, then as much detail as possible should be included in the main paper. 
        \item According to the NeurIPS Code of Ethics, workers involved in data collection, curation, or other labor should be paid at least the minimum wage in the country of the data collector. 
    \end{itemize}

\item {\bf Institutional review board (IRB) approvals or equivalent for research with human subjects}
    \item[] Question: Does the paper describe potential risks incurred by study participants, whether such risks were disclosed to the subjects, and whether Institutional Review Board (IRB) approvals (or an equivalent approval/review based on the requirements of your country or institution) were obtained?
    \item[] Answer: \answerNA{} 
    \item[] Guidelines:
    \begin{itemize}
        \item The answer \answerNA{} means that the paper does not involve crowdsourcing nor research with human subjects.
        \item Depending on the country in which research is conducted, IRB approval (or equivalent) may be required for any human subjects research. If you obtained IRB approval, you should clearly state this in the paper. 
        \item We recognize that the procedures for this may vary significantly between institutions and locations, and we expect authors to adhere to the NeurIPS Code of Ethics and the guidelines for their institution. 
        \item For initial submissions, do not include any information that would break anonymity (if applicable), such as the institution conducting the review.
    \end{itemize}

\item {\bf Declaration of LLM usage}
    \item[] Question: Does the paper describe the usage of LLMs if it is an important, original, or non-standard component of the core methods in this research? Note that if the LLM is used only for writing, editing, or formatting purposes and does \emph{not} impact the core methodology, scientific rigor, or originality of the research, declaration is not required.
    \item[] Answer: \answerNo{} 
    \item[] Guidelines:
    \begin{itemize}
        \item The answer \answerNA{} means that the core method development in this research does not involve LLMs as any important, original, or non-standard components.
        \item Please refer to our LLM policy in the NeurIPS handbook for what should or should not be described.
    \end{itemize}

\end{enumerate}

%% file: ref.bib
@article{shannon1948mathematical,
  title={A mathematical theory of communication},
  author={Shannon, Claude E},
  journal={The Bell System Technical Journal},
  volume={27},
  number={3},
  pages={379--423},
  year={1948}
}

@article{raiaan2024review,
  title={A review on large language models: Architectures, applications, taxonomies, open issues and challenges},
  author={Raiaan, Mohaimenul Azam Khan and Mukta, Md Saddam Hossain and Fatema, Kaniz and Fahad, Nur Mohammad and Sakib, Sadman and Mim, Most Marufatul Jannat and Ahmad, Jubaer and Ali, Mohammed Eunus and Azam, Sami},
  journal={IEEE access},
  volume={12},
  pages={26839--26874},
  year={2024},
  publisher={IEEE}
}

@article{naveed2025comprehensive,
  title={A comprehensive overview of large language models},
  author={Naveed, Humza and Khan, Asad Ullah and Qiu, Shi and Saqib, Muhammad and Anwar, Saeed and Usman, Muhammad and Akhtar, Naveed and Barnes, Nick and Mian, Ajmal},
  journal={ACM Transactions on Intelligent Systems and Technology},
  volume={16},
  number={5},
  pages={1--72},
  year={2025},
  publisher={ACM New York, NY}
}

@article{hagos2024recent,
  title={Recent advances in generative ai and large language models: Current status, challenges, and perspectives},
  author={Hagos, Desta Haileselassie and Battle, Rick and Rawat, Danda B},
  journal={IEEE transactions on artificial intelligence},
  volume={5},
  number={12},
  pages={5873--5893},
  year={2024},
  publisher={IEEE}
}

@inproceedings{chkirbene2024large,
  title={Large language models (llm) in industry: A survey of applications, challenges, and trends},
  author={Chkirbene, Zina and Hamila, Ridha and Gouissem, Ala and Devrim, Unal},
  booktitle={2024 IEEE 21st International Conference on Smart Communities: Improving Quality of Life using AI, Robotics and IoT (HONET)},
  pages={229--234},
  year={2024},
  organization={IEEE}
}

@inproceedings{miao2024demystifying,
  title={Demystifying data management for large language models},
  author={Miao, Xupeng and Jia, Zhihao and Cui, Bin},
  booktitle={Companion of the 2024 International Conference on Management of Data},
  pages={547--555},
  year={2024}
}

@inproceedings{zhang2024applications,
  title={Applications and challenges for large language models: From data management perspective},
  author={Zhang, Meihui and Ji, Zhaoxuan and Luo, Zhaojing and Wu, Yuncheng and Chai, Chengliang},
  booktitle={2024 IEEE 40th International Conference on Data Engineering (ICDE)},
  pages={5530--5541},
  year={2024},
  organization={IEEE}
}

@article{bagui2026large,
  title={Large Language Models in Database Management System Optimization: A Survey},
  author={Bagui, Sikha and Malagutti, Marta and Morelli, Riccardo and Tamascelli, Michael},
  journal={ACM Transactions on Intelligent Systems and Technology},
  year={2026},
  publisher={ACM New York, NY}
}

@inproceedings{tummuri2025enhancing,
  title={Enhancing Data Pipelines with Foundation Models: A New Approach to Automated Schema Mapping and SQL Generation},
  author={Tummuri, Sai Sukesh Reddy},
  booktitle={International Conference on Advances in Computer Engineering and Communication Systems},
  pages={21--30},
  year={2025},
  organization={Springer}
}

@article{chen2026large,
  title={Large language models for lossless image compression: Next-pixel prediction in language space is all you need},
  author={Chen, Kecheng and Zhang, Pingping and Liu, Hui and Liu, Jie and Liu, Yibing and Huang, Jiaxin and Wang, Shiqi and Yan, Hong and Li, Haoliang},
  journal={Advances in Neural Information Processing Systems},
  volume={38},
  pages={157548--157569},
  year={2026}
}

@inproceedings{kipf2026waiting,
  title={Waiting to Decompress: The Economics of LLM-Based Compression.},
  author={Kipf, Andreas and Schmidt, Tobias and Kuo, Ping-Lin and Krid, Skander and Rengert, Moritz and Heller, Luca and Zimmerer, Andreas and Stoian, Mihail and Pandey, Varun and van Renen, Alexander},
  booktitle={CIDR},
  year={2026}
}

@inproceedings{haddadi2026comparing,
  title={Comparing Text Compression Capabilities of Large Language Models with Traditional Compression Algorithms},
  author={Haddadi, Mehran and Teahan, William John},
  booktitle={Proceedings of the 19th Conference of the European Chapter of the Association for Computational Linguistics (Volume 4: Student Research Workshop)},
  pages={219--232},
  year={2026}
}

@article{witten1987arithmetic,
  author  = {Witten, Ian H. and Neal, Radford M. and Cleary, John G.},
  title   = {Arithmetic Coding for Data Compression},
  journal = {Communications of the ACM},
  volume  = {30},
  number  = {6},
  pages   = {520--540},
  year    = {1987}
}

@inproceedings{zhang2025l3tc,
  title={L3TC: Leveraging RWKV for learned lossless low-complexity text compression},
  author={Zhang, Junxuan and Cheng, Zhengxue and Zhao, Yan and Wang, Shihao and Zhou, Dajiang and Lu, Guo and Song, Li},
  booktitle={Proceedings of the AAAI Conference on Artificial Intelligence},
  volume={39},
  number={12},
  pages={13251--13259},
  year={2025}
}

@article{gloeckle2024better,
  title={Better \& faster large language models via multi-token prediction},
  author={Gloeckle, Fabian and Idrissi, Badr Youbi and Rozi{\`e}re, Baptiste and Lopez-Paz, David and Synnaeve, Gabriel},
  journal={arXiv preprint arXiv:2404.19737},
  year={2024}
}

@article{hu2025speculative,
  title={Speculative decoding and beyond: An in-depth survey of techniques},
  author={Hu, Yunhai and Liu, Zining and Dong, Zhenyuan and Peng, Tianfan and McDanel, Bradley and Zhang, Sai Qian},
  journal={arXiv preprint arXiv:2502.19732},
  year={2025}
}

@article{fu2025bits,
  title={From bits to rounds: Parallel decoding with exploration for diffusion language models},
  author={Fu, Hengyu and Huang, Baihe and Adams, Virginia and Wang, Charles and Srinivasan, Venkat and Jiao, Jiantao},
  journal={arXiv preprint arXiv:2511.21103},
  year={2025}
}

@article{chauhan2024brief,
  title={A brief review of hypernetworks in deep learning: VK Chauhan et al.},
  author={Chauhan, Vinod Kumar and Zhou, Jiandong and Lu, Ping and Molaei, Soheila and Clifton, David A},
  journal={Artificial Intelligence Review},
  volume={57},
  number={9},
  pages={250},
  year={2024},
  publisher={Springer}
}

@inproceedings{tan2026instant,
  title={Instant personalized large language model adaptation via hypernetwork},
  author={Tan, Zhaoxuan and Zhang, Zixuan and Wen, Haoyang and Li, Zheng and Zhang, Rongzhi and Chen, Pei and Mo, Fengran and Liu, Zheyuan and Zeng, Qingkai and Yin, Qingyu and others},
  booktitle={Proceedings of the 64th Annual Meeting of the Association for Computational Linguistics (Volume 1: Long Papers)},
  pages={23557--23580},
  year={2026}
}

@article{hu2021lora,
  title={Lora: Low-rank adaptation of large language models},
  author={Hu, Edward J and Shen, Yelong and Wallis, Phillip and Allen-Zhu, Zeyuan and Li, Yuanzhi and Wang, Shean and Wang, Lu and Chen, Weizhu},
  journal={arXiv preprint arXiv:2106.09685},
  year={2021}
}

@inproceedings{ha2017hypernetworks,
  title={HyperNetworks},
  author={Ha, David and Dai, Andrew and Le, Quoc V.},
  booktitle={International Conference on Learning Representations (ICLR)},
  year={2017}
}

@phdthesis{pasco1977source,
  author  = {Pasco, Richard C.},
  title   = {Source Coding Algorithms for Fast Data Compression},
  school  = {Stanford University},
  year    = {1977}
}

@article{rissanen1976generalized,
  author  = {Rissanen, Jorma J.},
  title   = {Generalized {K}raft Inequality and Arithmetic Coding},
  journal = {IBM Journal of Research and Development},
  volume  = {20},
  number  = {3},
  pages   = {198--203},
  year    = {1976},
  publisher = {IBM}
}

@inproceedings{deletang2024language,
  title     = {Language Modeling Is Compression},
  author    = {Del{\'e}tang, Gr{\'e}goire and Ruoss, Anian and
               Duquenne, Paul-Ambroise and Catt, Elliot and
               Genewein, Tim and Mattern, Christopher and
               Grau-Moya, Jordi and Wenliang, Li Kevin and
               Aitchison, Matthew and Orseau, Laurent and
               Hutter, Marcus and Veness, Joel},
  booktitle = {International Conference on Learning Representations},
  year      = {2024},
  url       = {https://proceedings.iclr.cc/paper_files/paper/2024/hash/3cbf627fa24fb6cb576e04e689b9428b-Abstract-Conference.html}
}

@inproceedings{chen2025large,
  title     = {Large Language Models for Lossless Image Compression:
               Next-Pixel Prediction in Language Space Is All You Need},
  author    = {Chen, Kecheng and Zhang, Pingping and Liu, Hui and
               Liu, Jie and Liu, Yibing and Huang, Jiaxin and
               Wang, Shiqi and Yan, Hong and Li, Haoliang},
  booktitle = {Advances in Neural Information Processing Systems},
  volume    = {38},
  year      = {2025},
  doi       = {10.52202/085713-5264},
  url       = {https://proceedings.neurips.cc/paper_files/paper/2025/hash/e7349e785900b93d8b4971a3f2c1cefe-Abstract-Conference.html}
}

@misc{zhao2026omnizip,
  title         = {{OmniZip}: Learning a Unified and Lightweight Lossless
                   Compressor for Multi-Modal Data},
  author        = {Zhao, Yan and Cheng, Zhengxue and Zhang, Junxuan and
                   Zhou, Dajiang and Gu, Qunshan and Wang, Qi and Song, Li},
  year          = {2026},
  eprint        = {2602.22286},
  archivePrefix = {arXiv},
  primaryClass  = {cs.LG},
  doi           = {10.48550/arXiv.2602.22286},
  url           = {https://arxiv.org/abs/2602.22286}
}

@article{li2025lossless,
  title   = {Lossless Data Compression by Large Models},
  author  = {Li, Ziguang and Huang, Chao and Wang, Xuliang and
             Hu, Haibo and Wyeth, Cole and Bu, Dongbo and Yu, Quan and
             Gao, Wen and Liu, Xingwu and Li, Ming},
  journal = {Nature Machine Intelligence},
  year    = {2025},
  doi     = {10.1038/s42256-025-01033-7},
  url     = {https://doi.org/10.1038/s42256-025-01033-7}
}

@misc{valmeekam2023llmzip,
  title         = {{LLMZip}: Lossless Text Compression Using Large
                   Language Models},
  author        = {Valmeekam, Chandra Shekhara Kaushik and
                   Narayanan, Krishna and Kalathil, Dileep and
                   Chamberland, Jean-Francois and Shakkottai, Srinivas},
  year          = {2023},
  eprint        = {2306.04050},
  archivePrefix = {arXiv},
  primaryClass  = {cs.IT},
  doi           = {10.48550/arXiv.2306.04050},
  url           = {https://arxiv.org/abs/2306.04050}
}

@misc{mittu2024finezip,
  title         = {{FineZip}: Pushing the Limits of Large Language Models
                   for Practical Lossless Text Compression},
  author        = {Mittu, Fazal and Bu, Yihuan and Gupta, Akshat and
                   Devireddy, Ashok and Ozdarendeli, Alp Eren and
                   Singh, Anant and Anumanchipalli, Gopala},
  year          = {2024},
  eprint        = {2409.17141},
  archivePrefix = {arXiv},
  primaryClass  = {cs.CL},
  doi           = {10.48550/arXiv.2409.17141},
  url           = {https://arxiv.org/abs/2409.17141}
}

@misc{heurteldepeiges2024compression,
  title         = {Compression via Pre-trained Transformers:
                   A Study on Byte-Level Multimodal Data},
  author        = {Heurtel-Depeiges, David and Ruoss, Anian and
                   Veness, Joel and Genewein, Tim},
  year          = {2024},
  eprint        = {2410.05078},
  archivePrefix = {arXiv},
  primaryClass  = {cs.LG},
  doi           = {10.48550/arXiv.2410.05078},
  url           = {https://arxiv.org/abs/2410.05078}
}

@misc{zhang2024l3tc,
  title         = {{L3TC}: Leveraging {RWKV} for Learned Lossless
                   Low-Complexity Text Compression},
  author        = {Zhang, Junxuan and Cheng, Zhengxue and Zhao, Yan and
                   Wang, Shihao and Zhou, Dajiang and Lu, Guo and Song, Li},
  year          = {2024},
  eprint        = {2412.16642},
  archivePrefix = {arXiv},
  primaryClass  = {cs.CL},
  doi           = {10.48550/arXiv.2412.16642},
  url           = {https://arxiv.org/abs/2412.16642}
}

@inproceedings{zhang2025testtime,
  title     = {Test-Time Steering for Lossless Text Compression via
               Weighted Product of Experts},
  author    = {Zhang, Qihang and Li, Muchen and Wang, Ziao and
               Liao, Renjie and Wang, Lele},
  booktitle = {Findings of the Association for Computational Linguistics:
               EMNLP 2025},
  pages     = {2076--2088},
  address   = {Suzhou, China},
  publisher = {Association for Computational Linguistics},
  month     = nov,
  year      = {2025},
  doi       = {10.18653/v1/2025.findings-emnlp.110},
  url       = {https://aclanthology.org/2025.findings-emnlp.110/}
}

@inproceedings{pan2025understanding,
  title     = {Understanding {LLM} Behaviors via Compression:
               Data Generation, Knowledge Acquisition and Scaling Laws},
  author    = {Pan, Zhixuan and Wang, Shaowen and Liao, Pengfei and Li, Jian},
  booktitle = {Advances in Neural Information Processing Systems},
  volume    = {38},
  year      = {2025},
  doi       = {10.52202/085713-5613},
  url       = {https://proceedings.neurips.cc/paper_files/paper/2025/file/f61d7778e89b9221d1ea0ce8428b7014-Paper-Conference.pdf}
}

@techreport{deutsch1996gzip,
  author      = {L. Peter Deutsch},
  title       = {{GZIP} File Format Specification Version 4.3},
  institution = {Internet Engineering Task Force},
  type        = {RFC},
  number      = {1952},
  year        = {1996},
  doi         = {10.17487/RFC1952}
}

@article{yang2023introduction,
  title     = {An Introduction to Neural Data Compression},
  author    = {Yang, Yibo and Mandt, Stephan and Theis, Lucas},
  journal   = {Foundations and Trends in Computer Graphics and Vision},
  volume    = {15},
  number    = {2},
  pages     = {113--200},
  year      = {2023},
  publisher = {Now Publishers},
  doi       = {10.1561/0600000107}
}

@inproceedings{austin2021structured,
  title     = {Structured Denoising Diffusion Models in Discrete State-Spaces},
  author    = {Austin, Jacob and Johnson, Daniel D. and Ho, Jonathan
               and Tarlow, Daniel and van den Berg, Rianne},
  booktitle = {Advances in Neural Information Processing Systems},
  volume    = {34},
  year      = {2021}
}

@inproceedings{hoogeboom2021argmax,
  title     = {Argmax Flows and Multinomial Diffusion:
               Learning Categorical Distributions},
  author    = {Hoogeboom, Emiel and Nielsen, Didrik and Jaini, Priyank
               and Forr{\'e}, Patrick and Welling, Max},
  booktitle = {Advances in Neural Information Processing Systems},
  volume    = {34},
  year      = {2021}
}

@inproceedings{arriola2025block,
  title     = {Block Diffusion: Interpolating Between Autoregressive
               and Diffusion Language Models},
  author    = {Arriola, Marianne and Gokaslan, Aaron and Chiu, Justin T.
               and Yang, Zhihan and Qi, Zhixuan and Han, Jiaqi
               and Sahoo, Subham Sekhar and Kuleshov, Volodymyr},
  booktitle = {International Conference on Learning Representations},
  year      = {2025}
}

@inproceedings{hu2022lora,
  title     = {{LoRA}: Low-Rank Adaptation of Large Language Models},
  author    = {Hu, Edward J. and Shen, Yelong and Wallis, Phillip
               and Allen-Zhu, Zeyuan and Li, Yuanzhi and Wang, Shean
               and Wang, Lu and Chen, Weizhu},
  booktitle = {International Conference on Learning Representations},
  year      = {2022}
}

@inproceedings{charakorn2025texttolora,
  title     = {Text-to-LoRA: Instant Transformer Adaptation},
  author    = {Charakorn, Rujikorn and Cetin, Edoardo and
               Tang, Yucheng and Lange, Robert Tjarko},
  booktitle = {Proceedings of the 42nd International Conference
               on Machine Learning},
  year      = {2025}
}

@article{charakorn2026doctolora,
  title   = {Doc-to-LoRA: Learning to Instantly Internalize Contexts},
  author  = {Charakorn, Rujikorn and Cetin, Edoardo and
             Uesaka, Shinnosuke and Lange, Robert Tjarko},
  journal = {arXiv preprint arXiv:2602.15902},
  year    = {2026}
}

@article{li2025eagle3,
  title   = {{EAGLE-3}: Scaling Up Inference Acceleration of Large Language Models via Training-Time Test},
  author  = {Li, Yuhui and Wei, Fangyun and Zhang, Chao and Zhang, Hongyang},
  journal = {arXiv preprint arXiv:2503.01840},
  year    = {2025},
  doi     = {10.48550/arXiv.2503.01840}
}

@article{chen2026dflash,
  title   = {{DFlash}: Block Diffusion for Flash Speculative Decoding},
  author  = {Chen, Jian and Liang, Yesheng and Liu, Zhijian},
  journal = {arXiv preprint arXiv:2602.06036},
  year    = {2026},
  doi     = {10.48550/arXiv.2602.06036}
}

@inproceedings{liu2025lmtp,
  title     = {{L-MTP}: Leap Multi-Token Prediction Beyond Adjacent Context for Large Language Models},
  author    = {Liu, Xiaohao and Xia, Xiaobo and Zhao, Weixiang and Zhang, Manyi and Yu, Xianzhi and Su, Xiu and Yang, Shuo and Ng, See-Kiong and Chua, Tat-Seng},
  booktitle = {Advances in Neural Information Processing Systems},
  year      = {2025}
}

@article{kirchenbauer2026multitoken,
  title   = {Multi-Token Prediction via Self-Distillation},
  author  = {Kirchenbauer, John and Hans, Abhimanyu and Bartoldson, Brian and Goldblum, Micah and Panda, Ashwinee and Goldstein, Tom},
  journal = {arXiv preprint arXiv:2602.06019},
  year    = {2026},
  doi     = {10.48550/arXiv.2602.06019}
}

@article{fu2026nemotrondiffusion,
  title   = {{Nemotron-Labs-Diffusion}: A Tri-Mode Language Model Unifying Autoregressive, Diffusion, and Self-Speculation Decoding},
  author  = {Fu, Yonggan and Whalen, Lexington and Garg, Abhinav and Wu, Chengyue and Khadkevich, Maksim and Oswald, Nicolai and Xie, Enze and Egert, Daniel and Sreenivas, Sharath Turuvekere and Diao, Shizhe and Yu, Chenhan and Yu, Ye and Chen, Weijia and Norouzi, Sajad and Liu, Jingyu and Lan, Shiyi and Zhu, Ligeng and Wang, Jin and Jiang, Jindong and Mardani, Morteza and Maghoumi, Mehran and Han, Song and Juki{\'c}, Ante and Tajbakhsh, Nima and Kautz, Jan and Molchanov, Pavlo},
  journal = {arXiv preprint arXiv:2607.05722},
  year    = {2026},
  doi     = {10.48550/arXiv.2607.05722}
}

@article{wu2025fastdllmv2,
  title   = {{Fast-dLLM v2}: Efficient Block-Diffusion LLM},
  author  = {Wu, Chengyue and Zhang, Hao and Xue, Shuchen and Diao, Shizhe and Fu, Yonggan and Liu, Zhijian and Molchanov, Pavlo and Luo, Ping and Han, Song and Xie, Enze},
  journal = {arXiv preprint arXiv:2509.26328},
  year    = {2025},
  doi     = {10.48550/arXiv.2509.26328}
}

@article{yang2025qwen3,
  title   = {{Qwen3} Technical Report},
  author  = {Yang, An and Li, Anfeng and Yang, Baosong and Zhang, Beichen and Hui, Binyuan and Zheng, Bo and Yu, Bowen and Gao, Chang and Huang, Chengen and Lv, Chenxu and Zheng, Chujie and Liu, Dayiheng and Zhou, Fan and Huang, Fei and Hu, Feng and Ge, Hao and Wei, Haoran and Lin, Huan and Tang, Jialong and Yang, Jian and Tu, Jianhong and Zhang, Jianwei and Yang, Jianxin and Yang, Jiaxi and Zhou, Jing and Zhou, Jingren and Lin, Junyang and Dang, Kai and Bao, Keqin and Yang, Kexin and Yu, Le and Deng, Lianghao and Li, Mei and Xue, Mingfeng and Li, Mingze and Zhang, Pei and Wang, Peng and Zhu, Qin and Men, Rui and Gao, Ruize and Liu, Shixuan and Luo, Shuang and Li, Tianhao and Tang, Tianyi and Yin, Wenbiao and Ren, Xingzhang and Wang, Xinyu and Zhang, Xinyu and Ren, Xuancheng and Fan, Yang and Su, Yang and Zhang, Yichang and Zhang, Yinger and Wan, Yu and Liu, Yuqiong and Wang, Zekun and Cui, Zeyu and Zhang, Zhenru and Zhou, Zhipeng and Qiu, Zihan},
  journal = {arXiv preprint arXiv:2505.09388},
  year    = {2025},
  doi     = {10.48550/arXiv.2505.09388}
}

@misc{hutter_enwik,
  author       = {Marcus Hutter},
  title        = {The Human Knowledge Compression Contest},
  howpublished = {\url{https://prize.hutter1.net/}},
  note         = {Source of the enwik8 dataset; accessed August 24, 2026},
  year         = {2006}
}

@inproceedings{leviathan2023fast,
  title     = {Fast Inference from Transformers via Speculative Decoding},
  author    = {Leviathan, Yaniv and Kalman, Matan and Matias, Yossi},
  booktitle = {Proceedings of the 40th International Conference on Machine Learning},
  pages     = {19274--19286},
  year      = {2023},
  volume    = {202},
  series    = {Proceedings of Machine Learning Research},
  publisher = {PMLR},
  url       = {https://proceedings.mlr.press/v202/leviathan23a.html}
}

@article{chen2023accelerating,
  title   = {Accelerating Large Language Model Decoding with Speculative Sampling},
  author  = {Chen, Charlie and Borgeaud, Sebastian and Irving, Geoffrey and
             Lespiau, Jean-Baptiste and Sifre, Laurent and Jumper, John},
  journal = {arXiv preprint arXiv:2302.01318},
  year    = {2023},
  doi     = {10.48550/arXiv.2302.01318},
  url     = {https://arxiv.org/abs/2302.01318}
}

@inproceedings{penedo2024fineweb,
  title     = {The FineWeb Datasets: Decanting the Web for the Finest Text Data at Scale},
  author    = {Penedo, Guilherme and Kydl{\'i}{\v{c}}ek, Hynek and
               Ben Allal, Loubna and Lozhkov, Anton and Mitchell, Margaret and
               Raffel, Colin and von Werra, Leandro and Wolf, Thomas},
  booktitle = {Advances in Neural Information Processing Systems},
  volume    = {37},
  pages     = {30811--30849},
  year      = {2024}
}

@article{charakorn2025text,
  title={Text-to-lora: Instant transformer adaption},
  author={Charakorn, Rujikorn and Cetin, Edoardo and Tang, Yujin and Lange, Robert Tjarko},
  journal={arXiv preprint arXiv:2506.06105},
  year={2025}
}

@article{zhang2025parallel,
  title   = {A Survey on Parallel Text Generation: From Parallel Decoding to Diffusion Language Models},
  author  = {Zhang, Lingzhe and Fang, Liancheng and Duan, Chiming and He, Minghua and Pan, Leyi and Xiao, Pei and Huang, Shiyu and Zhai, Yunpeng and Hu, Xuming and Yu, Philip S. and Liu, Aiwei},
  journal = {arXiv preprint arXiv:2508.08712},
  year    = {2025}
}

@article{xiang2026pmtp,
  title   = {{P-MTP}: Efficient Document Parsing via Multi-Token Prediction
             with Progressive Depth Scaling},
  author  = {Xiang, Le and Zhai, Chenxi and Wei, Shu and Wu, Jingjing
             and Xie, Qunyi and Tan, Xiao and Chen, Kunbin and He, Wei},
  journal = {arXiv preprint arXiv:2606.24447},
  year    = {2026},
  url     = {https://arxiv.org/abs/2606.24447}
}

@inproceedings{ivison2022hyperdecoders,
  title     = {{HyperDecoders}: Instance-Specific Decoders for Multi-Task {NLP}},
  author    = {Ivison, Hamish and Peters, Matthew},
  booktitle = {Findings of the Association for Computational Linguistics: EMNLP 2022},
  year      = {2022},
  pages     = {1715--1730},
  address   = {Abu Dhabi, United Arab Emirates},
  publisher = {Association for Computational Linguistics},
  doi       = {10.18653/v1/2022.findings-emnlp.124},
  url       = {https://aclanthology.org/2022.findings-emnlp.124/}
}

@article{yang2024qwen25,
  title   = {Qwen2.5 Technical Report},
  author  = {Yang, An and others},
  journal = {arXiv preprint arXiv:2412.15115},
  year    = {2024}
}

@techreport{bellard2021nncp,
  title       = {{NNCP} v2: Lossless Data Compression with Transformer},
  author      = {Bellard, Fabrice},
  institution = {Amarisoft},
  year        = {2021},
  month       = feb,
  url         = {https://bellard.org/nncp/nncp_v2.pdf}
}

@article{ren2026adapting,
  title   = {Adapting Diffusion Language Models for Lossless
             Pixel-Level Image Transmission},
  author  = {Ren, Tianqi and Li, Rongpeng and Chen, Xianfu and
             Li, Yingyu and Zhao, Zhifeng},
  journal = {arXiv preprint arXiv:2606.06273},
  year    = {2026},
  doi     = {10.48550/arXiv.2606.06273}
}

@inproceedings{shi2024simplified,
  title     = {Simplified and Generalized Masked Diffusion for
               Discrete Data},
  author    = {Shi, Jiaxin and Han, Kehang and Wang, Zhe and
               Doucet, Arnaud and Titsias, Michalis K.},
  booktitle = {Advances in Neural Information Processing Systems},
  volume    = {37},
  year      = {2024},
  doi       = {10.52202/079017-3277}
}

@article{nie2025llada,
  title   = {Large Language Diffusion Models},
  author  = {Nie, Shen and Zhu, Fengqi and You, Zebin and Zhang, Xiaolu and
             Ou, Jingyang and Hu, Jun and Zhou, Jun and Lin, Yankai and
             Wen, Ji-Rong and Li, Chongxuan},
  journal = {arXiv preprint arXiv:2502.09992},
  year    = {2025},
  doi     = {10.48550/arXiv.2502.09992}
}

@inproceedings{fu2026efficientdlm,
  title     = {{Efficient-DLM}: From Autoregressive to Diffusion
               Language Models, and Beyond in Speed},
  author    = {Fu, Yonggan and Whalen, Lexington and Ye, Zhifan and
               Dong, Xin and Diao, Shizhe and Liu, Jingyu and
               Wu, Chengyue and Zhang, Hao and Xie, Enze and Han, Song and
               Khadkevich, Maksim and Kautz, Jan and Lin, Yingyan Celine and
               Molchanov, Pavlo},
  booktitle = {Proceedings of the 43rd International Conference on
               Machine Learning},
  year      = {2026}
}

@article{liu2025sdlm,
  title   = {Sequential Diffusion Language Models},
  author  = {Liu, Yangzhou and Cao, Yue and Li, Hao and Luo, Gen and
             Chen, Zhe and Wang, Weiyun and Liang, Xiaobo and Qi, Biqing and
             Wu, Lijun and Tian, Changyao and Zhang, Yanting and
             Li, Yuqiang and Lu, Tong and Qiao, Yu and Dai, Jifeng and
             Wang, Wenhai},
  journal = {arXiv preprint arXiv:2509.24007},
  year    = {2025},
  doi     = {10.48550/arXiv.2509.24007}
}

@article{li2025prophet,
  title   = {Diffusion Language Models Know the Answer Before Decoding},
  author  = {Li, Pengxiang and Zhou, Yefan and Muhtar, Dilxat and
             Yin, Lu and Yan, Shilin and Shen, Li and Liang, Yi and
             Vosoughi, Soroush and Liu, Shiwei},
  journal = {arXiv preprint arXiv:2508.19982},
  year    = {2025},
  doi     = {10.48550/arXiv.2508.19982}
}

@inproceedings{mohamed2026sched,
  title     = {Fast-Decoding Diffusion Language Models via
               Progress-Aware Confidence Schedules},
  author    = {Mohamed, Amr and Zhang, Yang and Vazirgiannis, Michalis and
               Shang, Guokan},
  booktitle = {Findings of the Association for Computational Linguistics:
               ACL 2026},
  pages     = {35793--35807},
  year      = {2026},
  publisher = {Association for Computational Linguistics},
  doi       = {10.18653/v1/2026.findings-acl.1782}
}

@article{cai2026confidence,
  title   = {Confidence-Based Decoding Is Provably Efficient for
             Diffusion Language Models},
  author  = {Cai, Changxiao and Li, Gen},
  journal = {arXiv preprint arXiv:2603.22248},
  year    = {2026},
  doi     = {10.48550/arXiv.2603.22248}
}

@article{allal2025smollm2,
  title   = {{SmolLM2}: When Smol Goes Big---Data-Centric Training of a Small Language Model},
  author  = {Ben Allal, Loubna and Lozhkov, Anton and Bakouch, Elie and Mart{\'i}n Bl{\'a}zquez, Gabriel and Penedo, Guilherme and Tunstall, Lewis and Marafioti, Andr{\'e}s and Kydl{\'i}{\v{c}}ek, Hynek and Piqueres Lajar{\'i}n, Agust{\'i}n and Srivastav, Vaibhav and Lochner, Joshua and Fahlgren, Caleb and Nguyen, Xuan-Son and Fourrier, Cl{\'e}mentine and Burtenshaw, Ben and Larcher, Hugo and Zhao, Haojun and Zakka, Cyril and Morlon, Mathieu and Raffel, Colin and von Werra, Leandro and Wolf, Thomas},
  journal = {arXiv preprint arXiv:2502.02737},
  year    = {2025},
  doi     = {10.48550/arXiv.2502.02737},
  url     = {https://arxiv.org/abs/2502.02737}
}

@misc{dhankhar2026scaling,
      title={Scaling Laws for Hypernetwork-Based Knowledge Injection in Large Language Models}, 
      author={Nischay Dhankhar and Dos Baha and Abulhair Saparov},
      year={2026},
      eprint={2607.19604},
      archivePrefix={arXiv},
      primaryClass={cs.CL},
      url={https://arxiv.org/abs/2607.19604}, 
}
